\documentclass[lettersize,journal]{IEEEtran}
\usepackage{amsmath,amsfonts}
\usepackage{algorithmic}
\usepackage{algorithm}
\usepackage{array}
\usepackage[caption=false,font=normalsize,labelfont=sf,textfont=sf]{subfig}
\usepackage{textcomp}
\usepackage{stfloats}
\usepackage{url}
\usepackage{verbatim}
\usepackage{graphicx}
\usepackage{cite}
\usepackage{colortbl}
\usepackage{xcolor}
\usepackage{bm}
\usepackage{booktabs} 
\usepackage{multirow}
\usepackage{xspace}
\usepackage{pifont}
\newcommand{\cmark}{\ding{51}}
\definecolor{tabfirst}{rgb}{1.00, 0.90, 0.90}  
\definecolor{tabsecond}{rgb}{0.80, 0.95, 1.00} 
\definecolor{tabthird}{rgb}{1.00, 1.00, 0.85}  
\begin{document}

\title{D$^{3}$GS: Depth, DINO, and RGB Diffusion Co-Guided 3D Gaussian Splatting for Sparse-View Reconstruction}

\author{Yunqi Gao\textsuperscript{*},
Zhanfeng Liao\textsuperscript{*},
Hanzhang Tu,
Zhaoqi Su,
Guoqing Zheng,
Songtao Wang,
Hongwen Zhang,
Zhou Xue,
Leyuan Liu~\textsuperscript{$\dagger$},
Yebin Liu\textsuperscript{$\dagger$}~\IEEEmembership{Member,~IEEE}
\thanks{* Equal contribution.}
\thanks{$\dagger$ Corresponding authors.}%
\thanks{Yunqi Gao and Leyuan Liu are with Central China Normal University. E-mail: gaoyunqi@mails.ccnu.edu.cn, lyliu@mail.ccnu.edu.cn.}
\thanks{Zhanfeng Liao, Hanzhang Tu, and Yebin Liu are with Tsinghua University. E-mail:luckier003@gmail.com, thz22@mails.tsinghua.edu.cn, liuyebin@tsinghua.edu.cn }
\thanks{Zhaoqi Su is with Fuzhou University. E-mail: suzhaoqi@fzu.edu.cn.}
\thanks{Guoqing Zheng is with University of the Chinese Academy of Sciences. E-mail: zhengguoqing23@mails.ucas.ac.cn.}
\thanks{Songtao Wang is with Space Engineering University. E-mail:  wangsongtao1983@163.com }
\thanks{Hongwen Zhang is with Beijing Normal University. E-mail:  zhanghongwen@bnu.edu.cn.}
\thanks{Zhou Xue is with ByteDance Inc.. E-mail: xuezhou08@gmail.com.}
\thanks{Manuscript received April 19, 2021; revised August 16, 2021.}

}

\markboth{Journal of \LaTeX\ Class Files,~Vol.~14, No.~8, August~2021}%
{Shell \MakeLowercase{\textit{et al.}}: A Sample Article Using IEEEtran.cls for IEEE Journals}

\IEEEpubid{0000--0000/00\$00.00~\copyright~2021 IEEE}

\maketitle

\begin{abstract}
Novel view synthesis from sparse inputs remains challenging for 3D Gaussian Splatting (3DGS) due to ambiguous geometry, cross-view inconsistency, and missing details in under-constrained regions, resulting in degraded reconstruction and unstable rendering. To tackle these issues, we propose \textbf{D$^{3}$GS}, a \textbf{D}epth–\textbf{D}INO–\textbf{D}iffusion guided sparse-view Gaussian reconstruction framework that jointly enhances geometry and appearance.
\textbf{D$^{3}$GS} first recovers a high-resolution, metric depth map via diffusion-based completion and DPT (Dense Prediction Transformer) refinement, providing robust Gaussian initialization and geometric constraints. 
Then, a DINO-guided view-consistent learning is introduced to augment Gaussian attributes with structural features, improving multi-view consistency.
Finally, a diffusion-based Gaussian refinement module injects generative priors into an iterative optimization strategy, enhancing high-frequency geometric and appearance details within the Gaussian representation.
Experiments on DTU, LLFF, and Mip-NeRF 360 show that \textbf{D$^{3}$GS} achieves consistent and substantial improvements over strong baselines, with ablation studies validating the effectiveness and complementary roles of each component.
\end{abstract}


\section{Introduction}
 \label{sec:intro}
\IEEEPARstart{N}{ovel} view synthesis aims to render photorealistic images of a 3D scene from unseen viewpoints, a fundamental task in computer vision and graphics. Recent advances in 3DGS~\cite{3DGS} have significantly improved photo-realistic novel view synthesis by enabling efficient 3D scenes reconstruction with dense view supervision. However, their performance degrades notably in sparsely captured settings due to insufficient geometric and photometric supervision.
In such cases, the lack of reliable cross-view correspondence often leads to misaligned reconstructions and lost fine details, posing a major challenge for high-fidelity rendering under sparse inputs.
\begin{figure}[t]
    \centering
    \includegraphics[width=1.0\linewidth]{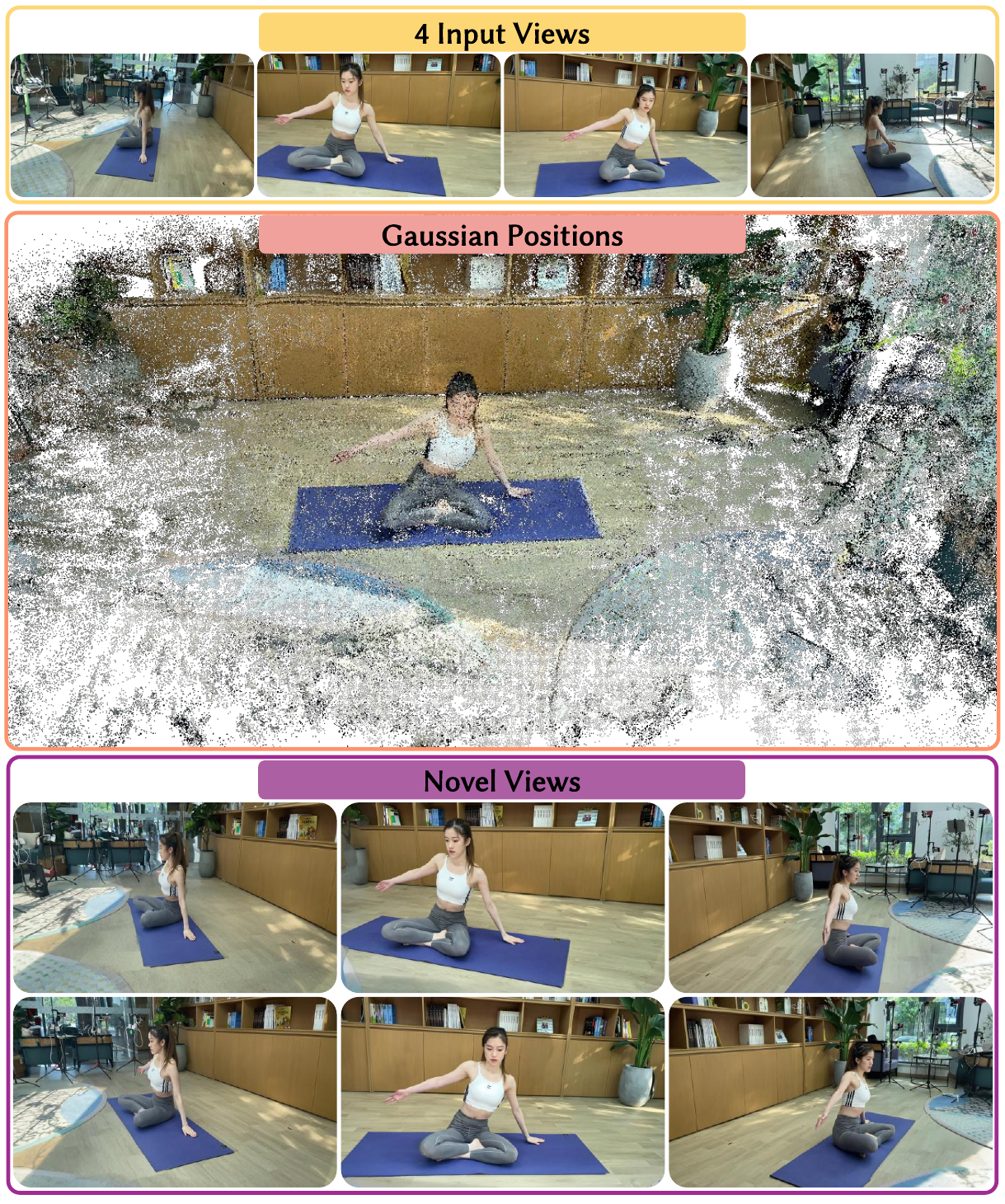}  
    \caption{Utilizing only four views covering an 180$^\circ$ scene, our method achieves accurate and coherent Gaussian primitives that enable clean geometric surfaces and minimal visual artifacts. Furthermore, the generated novel view renderings maintain strict view consistency and preserve fine high-frequency details, even in sparsely constrained regions.
    }
    \label{First_fig}
  
\end{figure}

\IEEEpubidadjcol
To mitigate the degradation in geometric reconstruction due to inadequate photometric supervision, prior studies~\cite{dngaussian,fsgs} have introduced depth regularization, which leverages structural priors from monocular depth estimators~\cite{depthanything}  to constrain geometry. However, view-consistent reconstruction remains challenging because of the scale ambiguity inherent in monocular depth. Some approaches~\cite{MAtCha,monofusion} aim to resolve this by aligning monocular depth~\cite{dust3r,MUSt3R} across input views, which improves results to some extent. However, since these depths are still not metric, absolute depth constraints cannot be applied, and floaters remain.
To address floaters in Gaussian reconstruction, some methods~\cite{difix3d,gsfixer,GenFusion} enhance reconstruction quality by integrating generative priors, such as pre-trained diffusion models~\cite{diffusion}, to supplement details in weakly supervised regions. However, due to the uncertain distribution of floaters, these methods generally demand large training datasets and sophisticated networks and when the 3D geometry is unreliable with initial reconstructions exhibiting cross-view inconsistencies, directly applying generative modules can further exacerbate these issues.

\begin{figure}[t]
    \centering
    \includegraphics[width=1.0\linewidth]{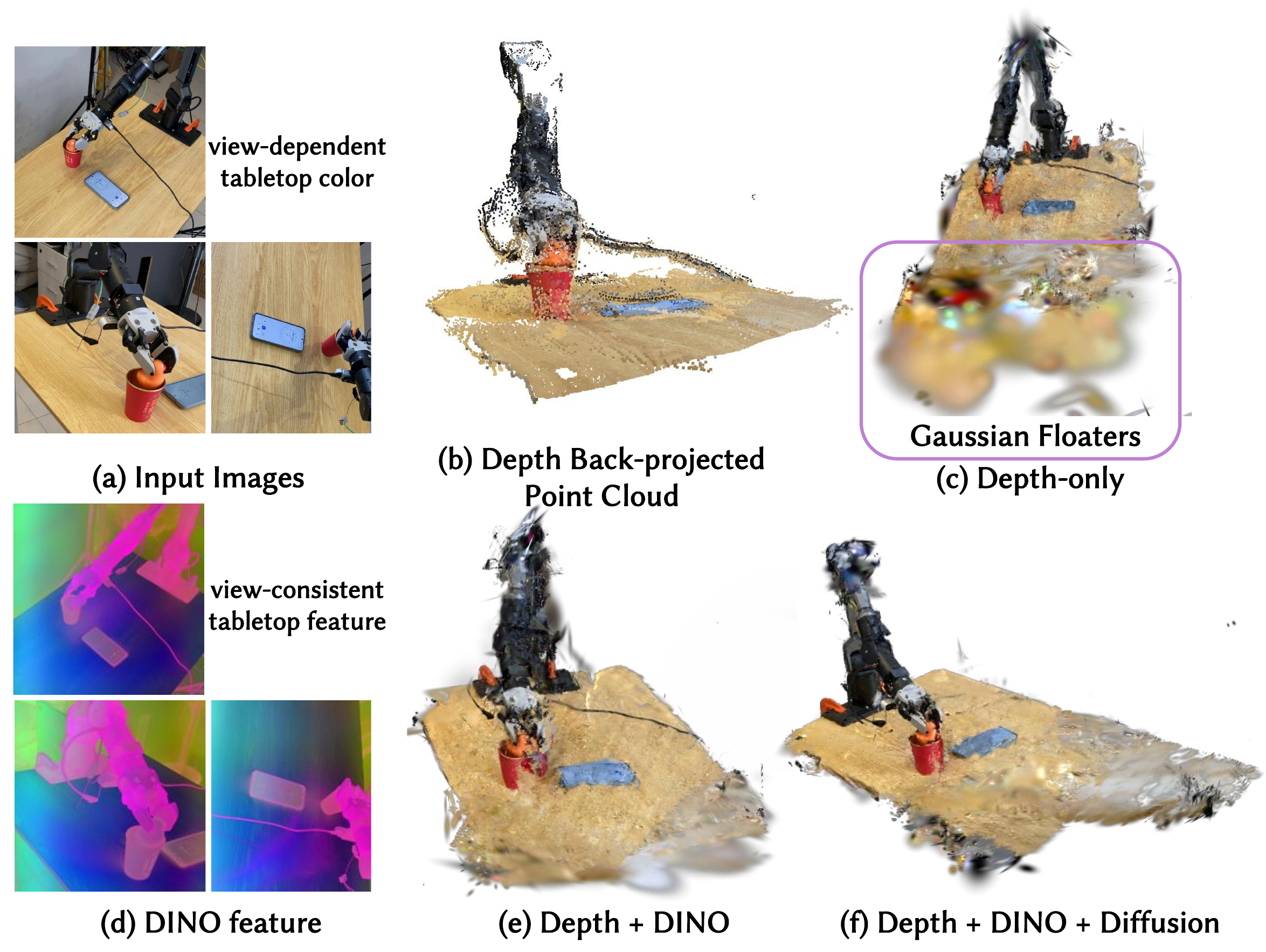}  
    \caption{The RGB appearance of the tabletop (a) varies significantly across viewpoints due to lighting changes, while (d) DINO features remain consistent across views. Although back-projection verifies the accuracy of the depth (b), cross-view color inconsistencies in the input images (a) still cause depth-only optimization to produce floating artifacts (c). Adding DINO supervision mitigates these floaters (e), and further introducing diffusion guidance improves fine details and visual completeness (f).
    }
    \label{fig:intro}
  
\end{figure}

In this paper, we revisit the problem of Gaussian reconstruction under sparse-view conditions. Existing methods~\cite{dngaussian,fsgs,MAtCha} typically rely on monocular or cross-view depth to alleviate geometric ambiguities, but these depths often suffer from scale inconsistency and noise, providing unreliable constraints and leading to floaters under sparse views. Therefore, introducing more accurate and scale-consistent metric depth is essential. Further observations show that even with relatively accurate depth, floaters still occur (Fig.~\ref{fig:intro}(b)(c)). This mainly stems from the volumetric rendering mechanism of 3DGS, where multiple Gaussians along the same ray contribute weighted effects, rather than producing a single intersection as in surface-based rendering. Under cross-view RGB inconsistencies, such as lighting variations, the competition among multiple Gaussians becomes unstable, weakening the effectiveness of depth constraints and introducing structural errors. These observations indicate that relying solely on RGB and depth constraints is insufficient for stable reconstruction. It is therefore necessary to incorporate cross-view consistent features to provide structural priors and enhance multi-view consistency (Fig.~\ref{fig:intro}(d)(e)). Moreover, due to the presence of unobserved regions caused by sparse inputs, details in these regions are difficult to recover when the viewpoint changes, leading to discontinuities or missing details in novel view rendering. To address this, generative strategies can be introduced to learn and restore details in unobserved regions (Fig.~\ref{fig:intro} (f)). Overall, reconstruction errors arise from the coupled under-constrained issues of insufficient geometric supervision, cross-view photometric inconsistencies, and visibility-induced detail loss.

To address this challenge, we propose $\mathbf{D^{3}GS}$, a sparse-view Gaussian Splatting framework that resolves geometric and appearance ambiguities by integrating metric depth, multi-view consistent features, and RGB diffusion-based generative priors.
For metric depth estimation, we design a low-to-high resolution strategy that leverages sparse metric depth from MapAnything~\cite{mapanything} and Bundle Adjustment~(BA)~\cite{ba}. Sparse metric depth cues are fed into a denoising U-Net~\cite{diffusion} to produce a globally consistent low-resolution depth, which is then refined by a DPT decoder~\cite{vit} to yield a high-resolution, scale-consistent metric depth. This depth map initializes Gaussian positions via reprojection and serves as a stable geometric constraint during optimization. 
To further improve multi-view consistency, we introduce a DINO-guided feature learning objective that complements RGB supervision. Specifically, each Gaussian is augmented with a compact DINO feature descriptor~\cite{simeoni2025dinov3}. Leveraging the inherent multi-view consistency of DINO features, this design provides a robust structural prior that enforces multi-view consistency and effectively mitigates floaters.
Finally, to recover high-frequency details in under-constrained regions, we introduce a diffusion-based novel-view refinement module. Unlike prior works that mainly use diffusion models for artifact removal, our approach focuses on enhancing fine-grained details and visual fidelity. We leverage a single-step diffusion model~\cite{diffusion}, fine-tuned to complete unseen regions and correct degraded renderings. To fully exploit the generative prior, we adopt an iterative optimization strategy that progressively refines novel views and feeds them back to update the Gaussian representation. By coupling appearance refinement with geometry optimization, the method achieves consistent improvements in structural accuracy and visual quality.

Extensive experiments on DTU~\cite{dtu}, LLFF~\cite{llff}, and Mip-NeRF 360~\cite{ReconFusion} datasets demonstrate consistent and significant improvements over strong baselines. Furthermore, as shown in Fig.~\ref{First_fig}, our method reconstructs stable and consistent scene geometry in large-scale scenes even with only four input views. Ablation studies further validate the effectiveness and complementary roles of each component within our framework. 
Our main contributions are summarized as follows:
\begin{itemize}

\item  We propose a novel Gaussian Splatting framework that resolves critical geometric and appearance ambiguities inherent in sparse inputs by integrating metric depth, DINO features, and RGB diffusion.

\item   We propose a cross-resolution depth refine module for obtaining high-resolution metric depth maps for robust 3D Gaussian optimization.

\item   We pioneer DINO-guided view-consistent feature learning for sparse-view reconstruction by augmenting Gaussian attributes with DINO feature descriptors, directly reinforcing the multi-view consistency and structural accuracy of the 3D Gaussian representation.

\item   We propose a novel-view diffusion-based iterative Gaussian refinement module for recovering fine details and enhancing visual fidelity in weakly-constrained regions, thereby further optimizing the Gaussian scene representation.
\end{itemize} 

\section{Related Works}
\label{sec:rel}
The field of scene reconstruction and novel view synthesis has been significantly advanced by seminal works such as NeRF~\cite{nerf} and 3DGS~\cite{3DGS}, inspiring a wide range of follow-up studies. In the following, we review representative methods most relevant to our work.

\subsection{Novel View Synthesis}
In recent years, Neural Radiance Fields (NeRF)~\cite{nerf} have made remarkable progress in encoding 3D scenes as implicit radiance fields. NeRF maps 3D coordinates and viewing directions to color and density values through a simple neural network, enabling novel view synthesis from a set of 2D images. Numerous variants have been proposed, with a particular focus on improving rendering quality~\cite{barron2021mip,barron2022mip} and accelerating training speed~\cite{fridovich2022plenoxels}. Despite these extensive efforts to enhance NeRF's capabilities, most existing methods~\cite{11455380} still suffer from time-consuming training and rendering processes, which limit their practicality in real-world applications.
To overcome these limitations, a new approach known as 3D Gaussian Splatting (3DGS)~\cite{3DGS} has recently emerged. 3DGS is an efficient method for novel view synthesis that reconstructs high-quality scenes by representing the radiance field with a set of 3D Gaussians. It performs particularly well on real-world data and is especially effective at capturing high-frequency details. Additionally, 3DGS demonstrates significant advantages in inference speed and supports more intuitive editing and interpretability compared to traditional neural implicit representations.
Motivated by the potential of 3DGS, recent studies have extended 3DGS~\cite{10521791} to improve rendering fidelity~\cite{yu24mipsplat,ye2024absgs,11455377,absgs}, enhance reconstruction quality~\cite{2dgs,pgsr,li2025geosplat,guedon2024sugar,gof,lyu20243dgsr}, and support 3D semantic segmentation via per-Gaussian feature embeddings~\cite{feature3dgs,saga,chen2025feat2gs}.

\subsection{Novel View Synthesis from Sparse Views}
With the emergence of 3DGS, several methods have explored its application in sparse-view novel view synthesis~\cite{fsgs,sparsegs,dngaussian,coherentgs,corgs,sparse2dgs,sharegs,fatesgs,monofusion,ni2025g4splat,11456202,zheng2025nexusgs,wan2025s2gaussian,wu2025sparse2dgs}. For instance, 
SparseGS~\cite{sparsegs} proposes to apply score distillation sampling loss~\cite{pooledreamfusion} to the training process for refining plausible details in regions with limited coverage in training views (i.e., sparse input views) and generating more complete 3D representations. CoherentGS~\cite{coherentgs} utilizes a pre-trained optical flow~\cite{shi2023flowformer++} model to regularize the pixel-wise correspondence between 3D Gaussians, which efficiently alleviates the overfitting problem by sparse input views.
DropGaussian~\cite{DropGaussian} randomly drops a subset of Gaussians during training (similar to dropout) to reduce overfitting to training views and improve novel view synthesis under sparse-view settings.
In addition to these approaches, other methods leverage depth-based regularization to guide Gaussian optimization for more accurate geometry.
FSGS~\cite{fsgs} and DNGaussian~\cite{dngaussian} employ depth constraints to guide the distribution of unstructured Gaussians along the scene surface, thereby improving the visual quality of novel views. These methods typically rely on depth priors obtained from pre-trained models~\cite{ranftl2021vision} as supervisory signals. However, such depth priors are often scale-ambiguous, especially under sparse-view conditions, making it difficult to achieve optimal Gaussian placement. MAtCha\cite{MAtCha} attempts to resolve this by injecting depth\cite{dust3r}, but struggles in non-overlapping or low-texture regions between input views. 
In contrast, we employ scale-aware depth supervision, using diffusion-completed metric depth seeded by sparse absolute-scale anchors to achieve more accurate, view-consistent Gaussian representation and higher-quality novel views.

\subsection{Generative Priors for Novel View Synthesis}

Generative models have demonstrated strong capability in inpainting plausible content in unobserved regions and restoring degraded areas with high visual fidelity. Recently, a series of works~\cite{ganerf,gsfixer,GenFusion,difix3d,ReconFusion,zhou2024diffgs} have leveraged generative priors~\cite{diffusion,li2023diffusion,ho2022videodiffusion,croitoru2023diffusion,yang2023diffusion2} to achieve significant progress in sparse-view novel view synthesis. 
Generative models have been leveraged to guide the optimization of NeRF~\cite{nerf} representations, such as ReconFusion~\cite{ReconFusion}, which combines image diffusion with PixelNeRF~\cite{yu2021pixelnerf}, and GANeRF~\cite{ganerf}, which incorporates GAN-based priors to enhance realism. More recent works~\cite{3dgsenhancer,difix3d,GenFusion,gsfixer} instead shift to 3D Gaussian~\cite{3DGS} representations, utilizing diffusion priors to enhance pseudo-observations rendered from the scene, thereby augmenting the training data and improving the optimization of the underlying 3D representation.

While our work follows a similar direction, we diverge in two key aspects.
(i) Existing approaches primarily focus on correcting artifacts in novel views, whereas we aim to further enhance fine-grained details and visual fidelity under already well-established geometry. (ii) We introduce an iterative 3D update scheme that establishes a closed-loop feedback between rendered views and the 3D representation, and achieves efficient refinement via single-step diffusion, thereby avoiding additional computational overhead despite the iterative optimization.

\section{Method}
\begin{figure*}[ht]
    \centering
    \includegraphics[width=1.0\linewidth]{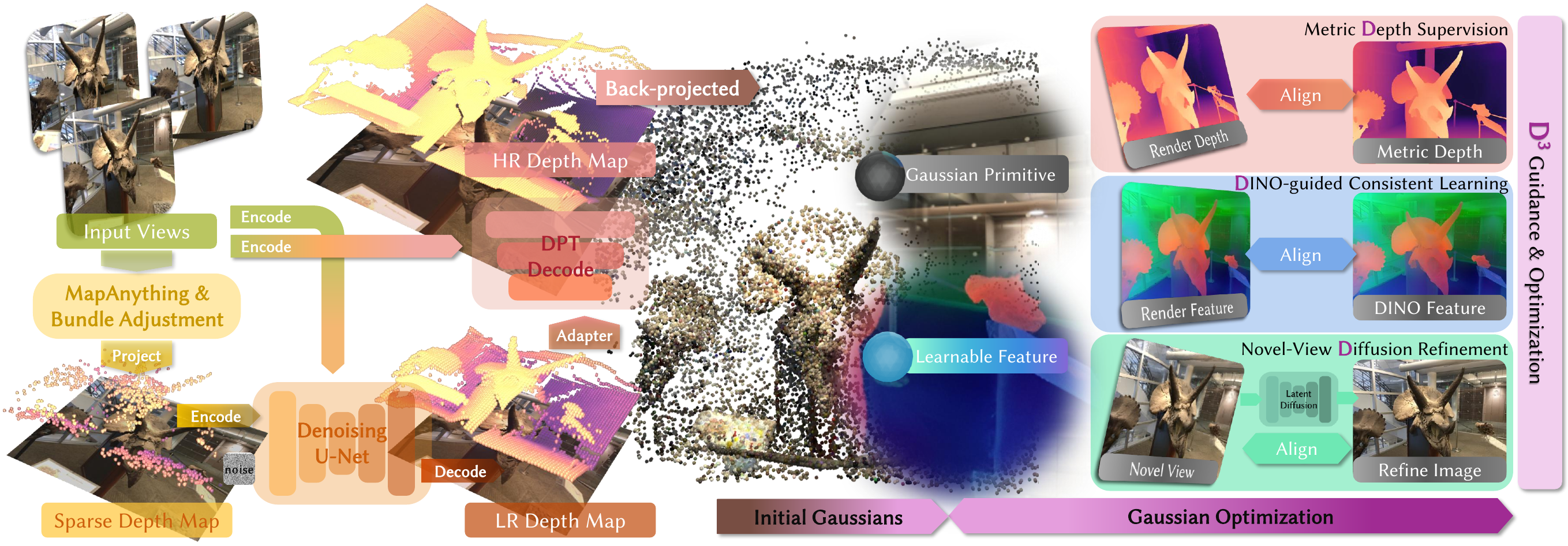}
    \caption{\textbf{Overview of D$^3$GS.}
We leverage metric sparse depth obtained from MapAnything and Bundle Adjustment as metric guidance to generate a complete high-resolution metric depth map via a denoising U-Net, followed by DPT-based refinement. The resulting depth map initializes Gaussian locations, upon which both the Gaussian locations and attributes are further optimized with three complementary strategies: (1) \emph{Metric depth supervision} enforces geometric consistency with the refined metric depth.
(2) \emph{DINO-guided view-consistent feature learning} augments Gaussian attributes with DINO descriptors to improve multi-view consistency.
(3) \emph{Diffusion-based novel-view refinement} iteratively enhances rendered results to recover high-frequency details in under-constrained regions.}
    \label{fig:framework}
\end{figure*}

Fig.~\ref{fig:framework} presents an overview of the pipeline. Given $N$ sparse input images, we aim to synthesize novel views with geometric fidelity and photorealistic detail. We introduce D$^3$GS, a sparse-view Gaussian reconstruction framework that couples metric depth guidance, DINO-guided multi-view consistency, and novel-view diffusion refinement within 3DGS optimization.
Sec.~\ref{sec:depth} details scale-consistent metric depth estimation.
Sec.~\ref{sec:dino} presents DINO-guided multi-view consistent learning.
Sec.~\ref{sec:diffusion} introduces diffusion-based Gaussian enhancement.
Sec.~\ref{sec:train_stay} summarizes the training strategy.
\subsection{Background}
\label{sec:method_core}

Gaussian splatting~\cite{3DGS} models a scene using 3D Gaussians defined as
\begin{equation}
    G_i(x) = e^{-\frac{1}{2}(x-\mu_i)^T\Sigma^{-1}(x-\mu_i)},
\end{equation}
where the covariance is parameterized as $\Sigma = RSS^TR^T$, with $S=(s_1,s_2,s_3)$ denoting axis-aligned scales and $R$ represented by a quaternion.
However, vanilla 3D Gaussians are not well aligned with underlying surfaces, which leads to suboptimal geometry and rendering quality under sparse views. 
To address this, we build upon surface-aligned, flattened Gaussians~\cite{pgsr} and further introduce a scale regularizer to stabilize optimization. Specifically, it (i) discourages anisotropy between the two dominant axes and (ii) prevents collapse of the smallest axis. 
\begin{equation}
\mathcal{L}_{\text{S}}(S)
=\lambda_{\text{iso}}\!\left|s_{\max}-s_{\text{mid}}\right|
+\lambda_{\text{bar}} \,|s_{\min}-\varepsilon|,
\label{loss_scale}
\end{equation}
where $(s_{\min},\, s_{\text{mid}},\, s_{\max})$ denote the sorted scales of $S$, and $\varepsilon$ prevents numerical collapse.

Each Gaussian also carries additional renderable and learnable attributes. To facilitate learning the multi-view consistent features, we augment the original Gaussian representation ${\{\mu_i,\, S_i,\, q_i,\, {o}_i,\, c_i\}}$, each Gaussian carries a learnable feature: 
\begin{equation}
    \Theta_i = \{\mu_i,\, S_i,\, q_i,\, o_i,\, c_i,\, f_i\},
    \label{eq:gs_param}
\end{equation}
where $\mu_i$ denotes the central location, $S_i$ the scaling vector, $q_i$ the rotation quaternion, ${o}_i$ the opacity, $c_i$ the color, and $f_i$ the feature vector. Notably, we augment the standard attributes with a 3-dimensional feature vector $ f_i \in \mathbb{R}^3$.

\noindent\textbf{Differentiable Depth Rasterization.}
We follow PGSR~\cite{pgsr} to perform differentiable depth rasterization. Specifically, surface normals are derived from the smallest Gaussian axis with view-consistent orientation, and both normals and plane distances are accumulated via $\alpha$-blending. The final depth is obtained by ray-plane intersection, which can be written as
\begin{equation}
    \bm{D}_{u}(\bm{p}) = 
    \frac{\sum_{i\in N} d_i \alpha_i \prod_{j<i}(1-\alpha_j)}
    {\left(\sum_{i\in N} \bm{n}_i \alpha_i \prod_{j<i}(1-\alpha_j)\right)^T 
    \bm{K}^{-1}\tilde{\bm{p}}}.
    \label{eq:depth}
\end{equation}
where $\bm{p}=[u,v]^T$ indicates the 2D position on the image plane, $d_i$ and $\bm{n}_i$ denote the distance and normal of the plane associated with the $i$-th Gaussian, $\alpha_i$ is its opacity, $\bm{K}$ is the camera intrinsic matrix, and $\tilde{\bm{p}}$ is the homogeneous coordinate of pixel $\bm{p}$.

\noindent\textbf{Differentiable Feature Rasterization.}
To obtain multi-view consistent features for supervision, we render the per-pixel feature map using the augmented Gaussian attributes. Each projected 2D Gaussian is sorted by depth, and the final feature is computed via $\alpha$-blending:
\begin{equation}
\bm{F} = \sum_{i=1}^{n} f_i\alpha_i
\prod_{j=1}^{i-1}(1-\alpha_j),
\label{gs:f}
\end{equation}
where $f_i$ denotes the feature of the $i$-th projected Gaussian, and $\alpha_i$ is the product of the projected 2D Gaussian density and the learned opacity value $o_i$ of the $i$-th Gaussian.

\noindent\textbf{Loss Function.} Our overall objective combines standard reconstruction losses, $\mathcal{L}_{\mathrm{I}}$ and $\mathcal{L}_{\mathrm{N}}$, with the three guidance losses $\mathcal{L}_{\text{de}}$, $\mathcal{L}_{\text{di}}$, and $\mathcal{L}_{\text{dif}}$, as well as the scale regularizer $\mathcal{L}_{\mathrm{S}}$:
\begingroup\small
\begin{equation}
\label{eq:total_loss}
\mathcal{L}_{\mathrm{total}}
=\mathcal{L}_{\text{S}}+\gamma_{\mathrm{i}}\mathcal{L}_{\mathrm{I}}
+\gamma_{\mathrm{n}}\mathcal{L}_{\text{N}}
+\gamma_{\mathrm{d}}\mathcal{L}_{\text{de}}
+\gamma_{\mathrm{e}}\mathcal{L}_{\text{di}}
+ m_t\,\gamma_{\mathrm{dif}}\mathcal{L}_{\text{dif}},\\
\end{equation}
\endgroup
where $\mathcal{L}_{\mathrm{I}}$ is RGB reconstruction loss, $\mathcal{L}_{\mathrm{N}}$ is normal consistency loss like ~\cite{pgsr}, $\mathcal{L}_{\text{de}}$ is scale-consistent metric depth loss, 
$\mathcal{L}_{\text{di}}$ is multi-view DINO feature alignment loss, 
$\mathcal{L}_{\text{dif}}$ is diffusion-based RGB refinement loss, 
and $m_t\!\in\!\{0,1\}$ is a curriculum mask for activating diffusion guidance.
\subsection{Metric Depth Estimation}
\label{sec:depth}

To address the geometric inaccuracies and the emergence of floating Gaussians inherent in sparse-view 3D reconstruction, we introduce an absolute metric-depth constraint that anchors reconstruction to the scene’s physical scale and improves cross-view consistency. Unlike prior methods that rely on relative depth~\cite{dngaussian,fsgs}, we estimate scale-consistent metric depth and use it as a grounded supervision signal.

To effectively inject multi-view information and alleviate the ill-posedness of monocular depth estimation, we introduce the SfM (Structure from Motion) point cloud as an explicit intermediate geometric representation to guide metric depth estimation. Compared to monocular depth that relies solely on image features, the SfM point cloud integrates multi-view observations and scene geometry, serving as a compact representation of multi-view structure and providing cross-view-consistent yet sparse geometric constraints. Moreover, SfM point clouds can be projected into each view according to visibility, forming image-aligned sparse depth maps that are naturally suitable for image-space depth estimation.
However, traditional SfM methods (e.g., COLMAP~\cite{colmap}) tend to degrade under sparse-view settings. To address this limitation, we adopt MapAnything~\cite{mapanything} to generate initial camera poses and point clouds, whose feed-forward architecture with learned priors enables more stable geometric estimation from sparse inputs. Based on this initialization, we further perform BA~\cite{ba} to jointly refine camera parameters and sparse 3D points by minimizing multi-view reprojection errors. Finally, the optimized point cloud is projected onto each view to obtain sparse depth maps.

\begin{figure}[t!]
    \centering
    \includegraphics[width=1.0\linewidth]{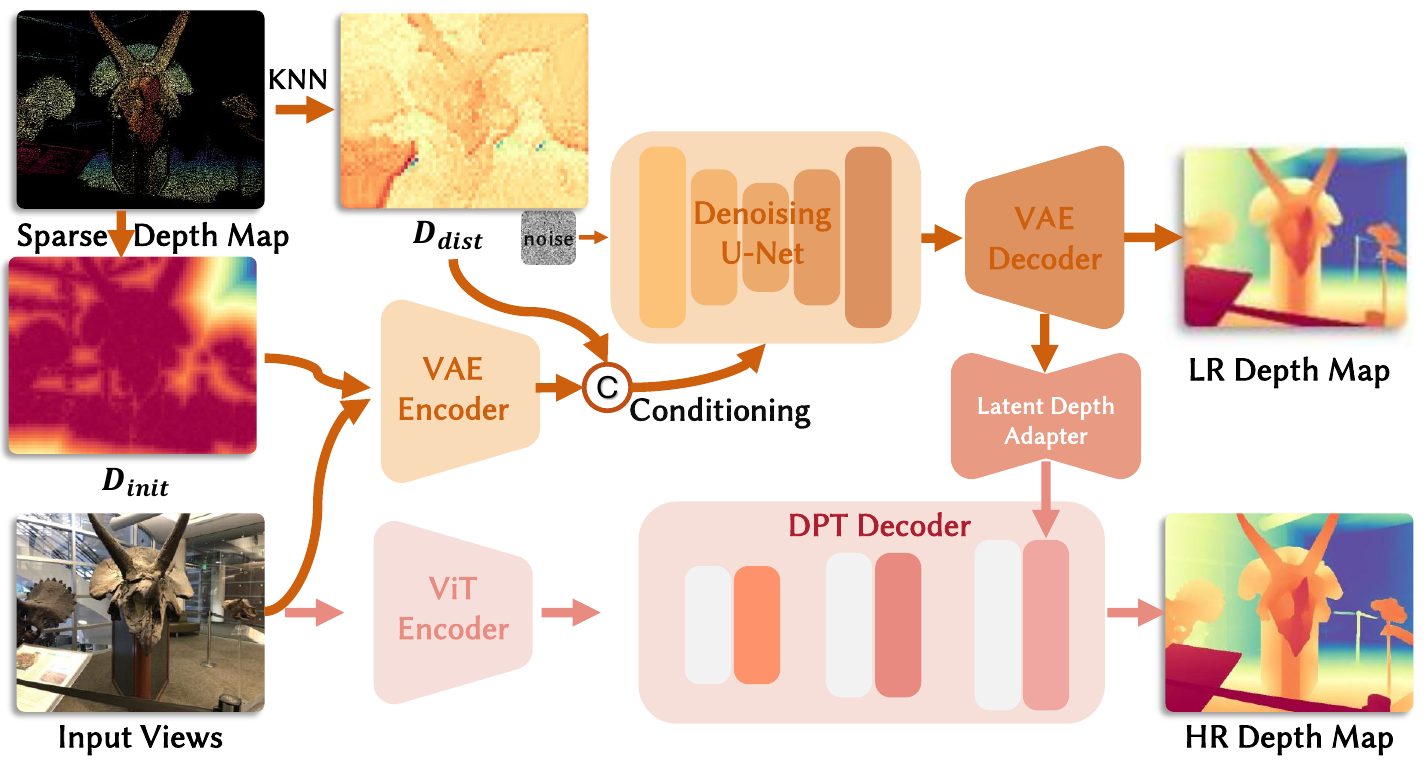}  
    \caption{\textbf{Metric depth estimation pipeline.}
The denoising U-Net first propagates sparse depth cues to estimate a globally consistent low-resolution metric depth map. This depth is then transformed via a lightweight adapter and injected into a DPT decoder, where it is fused with multi-scale image features to recover fine details and produce a high-resolution, scale-consistent metric depth.
    }
    \label{fig:metric_depth}
\end{figure}

To balance global scale accuracy and local detail fidelity, we design a depth module that progressively estimates metric depth from low to high resolution, as shown in Fig.~\ref{fig:metric_depth}.
Starting from sparse depth maps, we first obtain a dense depth initialization $\bm{D}_{\mathrm{init}}$ in the image domain via k-nearest neighbors (KNN) interpolation, and compute the per-pixel Euclidean distance-to-anchor transform $\bm{D}_{\mathrm{dist}}$.
These cues are then fused into a single latent-space condition by encoding the RGB image and $\bm{D}_{\mathrm{init}}$ with the VAE encoder $\mathcal{E}_{1}$~\cite{vae}, and injecting the latent-resolution $\bm{D}_{\mathrm{dist}}$:
\begin{equation}
\widetilde{\bm{D}}
= \mathcal{U}\!\Big(
  \mathcal{C}\big(
    \mathcal{E}_{1}(\bm{I}),\;
    \mathcal{E}_{1}\!\big(\mathcal{T}_{3}(\bm{D}_{\mathrm{init}})\big),\;
    \mathcal{M}(\bm{D}_{\mathrm{dist}})
  \big)
\Big),
\end{equation}
where $\mathcal{T}_{3}$ replicates a single channel into three channels, $\mathcal{M}$ resizes inputs to the VAE latent spatial resolution, and $\mathcal{C}$ denotes channel-wise concatenation. $\mathcal{U}$ is the denoising U\hbox{-}Net pretrained in~\cite{murre}, and $\widetilde{\bm{D}}\in[0,1]$ is the normalized depth map in image coordinates. The distance map $\bm{D}_{\mathrm{dist}}$ carries low-frequency geometry and is therefore not passed through the VAE encoder.

We map the normalized output back to the metric scale:
\begin{equation}
\bm{D}^{\text{lr}}(\bm{p}) = s \cdot [a + (b-a)\,\widetilde{\bm{D}}(\bm{p})] + t,
\end{equation}
where $a,b$ are the scene-scale parameters, $s$ and $t$ are the optimal scale and translation estimated robustly via RANSAC~\cite{fischler1981random}. $\bm{D}^{\mathrm{lr}}$ is the final aligned low-resolution metric depth.

The low-resolution metric depth $\bm{D}^{\mathrm{lr}}$ provides reliable global geometry, but it lacks fine spatial details due to its limited resolution. Instead of direct upsampling, we transform $\bm{D}^{\mathrm{lr}}$ into a feature-level geometric condition via a lightweight adapter, enabling scale alignment and compatibility with decoder features. 
We adopt a concise fusion architecture tailored to DPT-based depth foundation models, where a vision transformer extracts multi-scale image features, and the adapted depth features are injected into the decoder and fused with intermediate representations at multiple stages.
By aggregating image features with the injected geometric condition, the decoder produces a high-resolution, scale-consistent metric depth:
\begin{equation}
\bm{D}^{\mathrm{hr}} = \mathcal{D}_{1}\big(\mathcal{V}(\bm{I}), \mathcal{A}(\bm{D}^{\mathrm{lr}})\big),
\end{equation}
where $\mathcal{A}(\cdot)$ denotes the depth adapter, $\mathcal{D}_{\mathrm{1}}$ is the DPT decoder, $\mathcal{V}$ is the vision transformer~\cite{dinov2}, $\bm{I}$ is the input image. 
The low-to-high resolution geometric condition provides precise spatial cues, guiding the depth model to recover accurate local geometry and sharp structural details, resulting in high-resolution and scale-consistent metric depth.

we enforce an L1 consistency between $\bm{D}^{\mathrm{hr}}$ and the unbiased projected depth $\bm{D}_{u}$ (Eq.~\ref{eq:depth}) from the 3D Gaussians  :
\begin{equation}
\label{eq:d}
\mathcal{L}_{\text{de}}
= \frac{1}{Z}\sum_{\bm{p}\in\Omega}
\bigl|\,\bm{D}^{\mathrm{hr}}(\bm{p}) - \bm{D}_{u}(\bm{p})\,\bigr|,
\end{equation}
where $\bm{p}$ is pixels, $\Omega$ is the pixel domain and $Z=|\Omega|$ normalizes over pixels.

\subsection{DINO-guided Multi-view Consistent Learning}
\label{sec:dino}

Further observations show that even with accurate depth, floaters persist (Fig.~\ref{fig:intro}(b)(c)), due to the ambiguity of volumetric rendering under cross-view RGB inconsistencies. This indicates that geometric supervision alone is insufficient, and additional cross-view consistent cues are required for stable reconstruction.
To address this, we introduce multi-view consistent feature supervision to provide robust structural priors across views. An ideal feature representation should be invariant to appearance changes while preserving geometric correspondence. Recent advances in image foundation models have demonstrated strong 3D awareness~\cite{simeoni2025dinov3}, producing features that remain consistent across viewpoints for the same physical point.

Building upon this observation, we leverage the pretrained DINOv3~\cite{simeoni2025dinov3} backbone to extract multi-view consistent features. Prior works~\cite{saga,semanticgs,feature3dgs} typically rely on per-view segmentation maps, which often suffer from inconsistent predictions across views. In contrast, DINOv3 features exhibit strong geometric and semantic correspondence across viewpoints, enabling stable feature alignment without explicit supervision.

For each training image, we extract the features from the backbone. We then perform PCA~\cite{PCA} on features sampled from all training images and retain the top three components, yielding $\bm{F}_{gt}\!\in\!\mathbb{R}^{H\times W\times 3}$ for supervision.
To represent these features in the 3D Gaussians, each Gaussian $g_i\!\in\!\mathcal{G}$ is assigned a compact learnable feature ${f}_i\!\in\!\mathbb{R}^3$~(Eq.~\ref{eq:gs_param}). 
During training, the feature map $\bm{F}$ is rendered via differentiable feature rasterization~(Eq.~\ref{gs:f}). We supervise $\bm{F}$ with $\bm{F}_{gt}$:
\begin{equation}
\label{eq:feat_loss}
\mathcal{L}_{\text{di}}
=\frac{1}{Z}\sum_{\bm{p}\in\Omega}
\bigl\|\bm{F}(\bm{p})-\bm{F}_{gt}(\bm{p})\bigr\|_2^2,
\end{equation}
where $\bm{p}$ is pixels, $\Omega$ is the pixel domain and $Z=|\Omega|$ normalizes over pixels.

\subsection{Diffusion-based Gaussian Enhancement}
 \label{sec:diffusion}
Despite improved geometry and feature consistency, novel-view renderings still suffer from missing details and visual artifacts, especially in regions that are unobserved in sparse inputs. Such regions are inherently under-constrained, making it difficult to recover high-frequency details through geometric and photometric supervision alone, often leading to blurry textures or missing details.

To address this limitation, we introduce a generative prior to hallucinate plausible details in these under-constrained regions. Unlike prior work such as~\cite{difix3d,GenFusion,gsfixer}, which mainly applies diffusion models to suppress floaters or correct structural artifacts, our goal is to enhance fine-grained details and improve visual fidelity in novel-view synthesis. 
To this end, we adopt a single-step diffusion model~\cite{diffusion} to avoid the high computational cost of multi-step denoising, and iteratively refine details in novel views while feeding the enhanced results back into the Gaussian optimization process.
This is motivated by the observation that detail recovery and geometry refinement are inherently coupled, where improving appearance can further guide the optimization of the underlying 3D representation. 

Inspired by ~\cite{difix3d}, we leverage a single-step diffusion model $\mathcal{Q}_{\phi}$ as a generative prior to refine rendered novel views and guide the optimization of the explicit 3D Gaussian representation. To further enhance fine-grained details, we fine-tune $\mathcal{Q}_{\phi}$ on the SynCamMaster dataset~\cite{syncammaster}.
For each scene, we select 10 views, including 5 for training and 5 for testing. We first optimize the 3D Gaussians using the 5 training views and then render the 5 test views, which exhibit degraded quality with insufficient detail, blurred textures, and locally unseen regions.
The Gaussian optimization strictly follows our previous setup. All views are then used for fine-tuning, where the degraded render serves as the input, the reference images provide the conditioning, and the ground-truth image acts as the target view. For each target view, a subset of the remaining views from the same scene is randomly sampled as reference images $R_n$. 
After fine-tuning, the parameters $\phi$ are kept frozen during Gaussian optimization.

Given a coarse novel view $\bm{I}'_n$ and the reference image $R_n$, the refiner outputs an enhanced image
\begin{equation}
    \hat{\bm{I}}_n \;=\; \mathcal{Q}_{\phi}\!\left(\bm{I}'_n;\, R_n\right).
\end{equation}

To effectively integrate the generative prior into model training, we adopt an iterative optimization strategy that allows the generative model to progressively assist the Gaussian reconstruction process. Specifically, we first train the Gaussians for $T_1$ steps with only the original images as supervision, excluding the generative prior, to obtain a stable initial reconstruction. Following the warm-up, joint optimization with the generative prior proceeds, with generative refinement for every $T_2$ steps. At each update, given the current Gaussians $G_t$ and a target camera $v$, a novel view $\bm{I}'_n$ is rendered.
The Gaussians  can be optimized via the photometric loss between rendered images $\bm{I}'_n$ and enhanced image  $\hat {\bm{I}}_n$ by
\begingroup\small
\begin{equation}
\mathcal{L}_{\text{dif}}
= \lambda_{1}\,\mathcal{L}_{1}(\bm{I}'_n,\hat {\bm{I}}_n)
 + \lambda_{2}\,\mathcal{L}_{\mathrm{lp}}(\bm{I}'_n,\hat {\bm{I}}_n)
 + \lambda_{3}\,\mathcal{L}_{\mathrm{ss}}(\bm{I}'_n,\hat {\bm{I}}_n),
\end{equation}
\endgroup
where $\lambda_{1},\lambda_{2},\lambda_{3}\!$ are loss weights, $\mathcal{L}_{1}$ is L1 loss, $\mathcal{L}_{\mathrm{lp}}$ is the LPIPS loss, and $\mathcal{L}_{\mathrm{ss}}$ is the SSIM loss. We set $\lambda_{1}=0.8$, $\lambda_{2}=0.01$ and $\lambda_{3}=0.2$, respectively. 
This iterative refinement enables cross-view detail completion, enhances texture fidelity and edge sharpness, and effectively suppresses artifacts to improve structural consistency.

\subsection{Training Strategy}
\label{sec:train_stay}
Our training pipeline encompasses a continuous Gaussian optimization process designed to progressively improve coverage and cross-view consistency, ensuring the reconstruction starts from reliable geometry. The pipeline begins with Initialization, where we estimate the metric depth and extract DINO features (Sec.~\ref{sec:depth}, Sec.~\ref{sec:dino}). The Gaussian parameters are then initialized directly from the resulting point clouds, establishing a baseline 3D representation.

The iterative optimization adopts a step-wise constraint strategy. For the first $T_1$ iterations, Gaussians are optimized with only photometric supervision to quickly obtain a stable geometric foundation in visible regions. Afterwards, all three guidances (Depth, DINO, and Diffusion) are enabled, with the generative prior injected every $ T_2$ steps (Sec.~\ref{sec:diffusion}) to refine Gaussian attributes, improve scene coverage, and correct geometric misalignments.

\begin{table*}[h]

\centering
\caption{Quantitative comparison on LLFF and DTU. The top-3 results in each column are marked as \textbf{\colorbox{tabfirst}{1st}, \colorbox{tabsecond}{2nd}, \colorbox{tabthird}{3rd}}.}
\label{tab:llff_dtu_unified}
\renewcommand{\arraystretch}{1.12}
\setlength{\tabcolsep}{3.5pt}
\scriptsize
\resizebox{\textwidth}{!}{%
\begin{tabular}{l|ccc|ccc|ccc|ccc|ccc|ccc}
\toprule
\multirow{4}{*}{\textbf{Methods}} &
\multicolumn{9}{c|}{\textbf{LLFF}} &
\multicolumn{9}{c}{\textbf{DTU}} \\
\cmidrule(lr){2-10}\cmidrule(lr){11-19}
& \multicolumn{3}{c|}{\textbf{3-view}} & \multicolumn{3}{c|}{\textbf{6-view}} & \multicolumn{3}{c|}{\textbf{9-view}} &
\multicolumn{3}{c|}{\textbf{3-view}} & \multicolumn{3}{c|}{\textbf{6-view}} & \multicolumn{3}{c}{\textbf{9-view}} \\
\cmidrule(lr){2-4}\cmidrule(lr){5-7}\cmidrule(lr){8-10}\cmidrule(lr){11-13}\cmidrule(lr){14-16}\cmidrule(lr){17-19}
& PSNR$\uparrow$ & SSIM$\uparrow$ & LPIPS$\downarrow$
& PSNR$\uparrow$ & SSIM$\uparrow$ & LPIPS$\downarrow$
& PSNR$\uparrow$ & SSIM$\uparrow$ & LPIPS$\downarrow$
& PSNR$\uparrow$ & SSIM$\uparrow$ & LPIPS$\downarrow$
& PSNR$\uparrow$ & SSIM$\uparrow$ & LPIPS$\downarrow$
& PSNR$\uparrow$ & SSIM$\uparrow$ & LPIPS$\downarrow$ \\
\midrule
PGSR~\cite{pgsr}
& 12.52 & 0.302 & 0.512
& 16.93 & 0.570 & 0.314
& 18.21 & 0.613 & 0.278
& 13.77 & 0.794 & 0.215
& 22.96 & 0.902 & 0.101
& \cellcolor{tabthird}26.98 & \cellcolor{tabthird}0.947 & \cellcolor{tabthird}0.045 \\
DNGaussian~\cite{dngaussian}
& 19.12 & 0.591 & 0.294
& 22.18 & 0.755 & 0.198
& 23.17 & 0.788 & 0.180
& 18.91 & 0.790 & 0.176
& 22.10 & 0.851 & 0.148
& 23.94 & 0.887 & 0.131 \\
FSGS~\cite{fsgs}
& 20.31 & 0.652 & 0.288
& 24.20 & 0.811 & 0.173
& 25.32 & 0.856 & 0.136
& 17.34 & 0.818 & 0.169
& 21.55 & 0.880 & 0.127
& 24.33 & 0.911 & 0.106 \\
CoR-GS~\cite{corgs}
& 20.45 & 0.712 & \cellcolor{tabthird}0.196
& 24.49 & \cellcolor{tabthird}0.837 & \cellcolor{tabthird}0.115
& 26.06 & \cellcolor{tabthird}0.874 & \cellcolor{tabthird}0.089
& \cellcolor{tabthird}19.21 & \cellcolor{tabthird}0.853 & \cellcolor{tabthird}0.119
& \cellcolor{tabsecond}24.51 & \cellcolor{tabthird}0.917 & \cellcolor{tabsecond}0.068
& \cellcolor{tabfirst}27.18 & \cellcolor{tabthird}0.947 & \cellcolor{tabthird}0.045 \\
DropGaussian~\cite{DropGaussian}
& \cellcolor{tabthird}20.76 & \cellcolor{tabthird}0.713 & 0.200
& \cellcolor{tabthird}24.74 & \cellcolor{tabthird}0.837 & 0.117
& \cellcolor{tabsecond}26.21 & \cellcolor{tabthird}0.874 & \cellcolor{tabsecond}0.088
& 16.91 & 0.839 & 0.140
& 22.97 & \cellcolor{tabsecond}0.920 & \cellcolor{tabthird}0.073
& 26.17 & \cellcolor{tabsecond}0.952 & \cellcolor{tabsecond}0.044 \\
BinocularGS~\cite{Binoculargs}
& \cellcolor{tabsecond}21.44 & \cellcolor{tabsecond}0.754 & \cellcolor{tabsecond}0.168
& \cellcolor{tabsecond}24.87 & \cellcolor{tabsecond}0.845 & \cellcolor{tabsecond}0.106
& \cellcolor{tabthird}26.17 & \cellcolor{tabsecond}0.877 & 0.090
& \cellcolor{tabsecond}20.71 & \cellcolor{tabsecond}0.862 & \cellcolor{tabsecond}0.111
& \cellcolor{tabthird}24.31 & \cellcolor{tabthird}0.917 & \cellcolor{tabthird}0.073
& 26.70 & \cellcolor{tabthird}0.947 & 0.052 \\
AnySplat~\cite{anysplat}
& 14.87 & 0.633 & 0.281
& 16.82 & 0.681 & 0.221
& 18.24 & 0.474 & 0.315
& 17.25 & 0.733 & 0.182
& 20.66 & 0.776 & 0.142
& 21.72 & 0.792 & 0.130 \\
Difix3D+~\cite{difix3d}
& 19.23 & 0.608 & 0.244
& 21.89 & 0.707 & 0.188
& 22.51 & 0.730 & 0.175
& 17.22 & 0.773 & 0.168
& 21.26 & 0.825 & 0.127
& 23.72 & 0.857 & 0.105 \\
Ours
& \cellcolor{tabfirst}21.47 & \cellcolor{tabfirst}0.775 & \cellcolor{tabfirst}0.131
& \cellcolor{tabfirst}24.89 & \cellcolor{tabfirst}0.897 & \cellcolor{tabfirst}0.072
& \cellcolor{tabfirst}26.59 & \cellcolor{tabfirst}0.935 & \cellcolor{tabfirst}0.045
& \cellcolor{tabfirst}21.34 & \cellcolor{tabfirst}0.876 & \cellcolor{tabfirst}0.087
& \cellcolor{tabfirst}26.04 & \cellcolor{tabfirst}0.939 & \cellcolor{tabfirst}0.034
& \cellcolor{tabsecond}27.16 & \cellcolor{tabfirst}0.959 & \cellcolor{tabfirst}0.032 \\
\bottomrule
\end{tabular}
}%

\end{table*}

\begin{figure*}[h]
    \centering
    \includegraphics[width=0.98\linewidth]{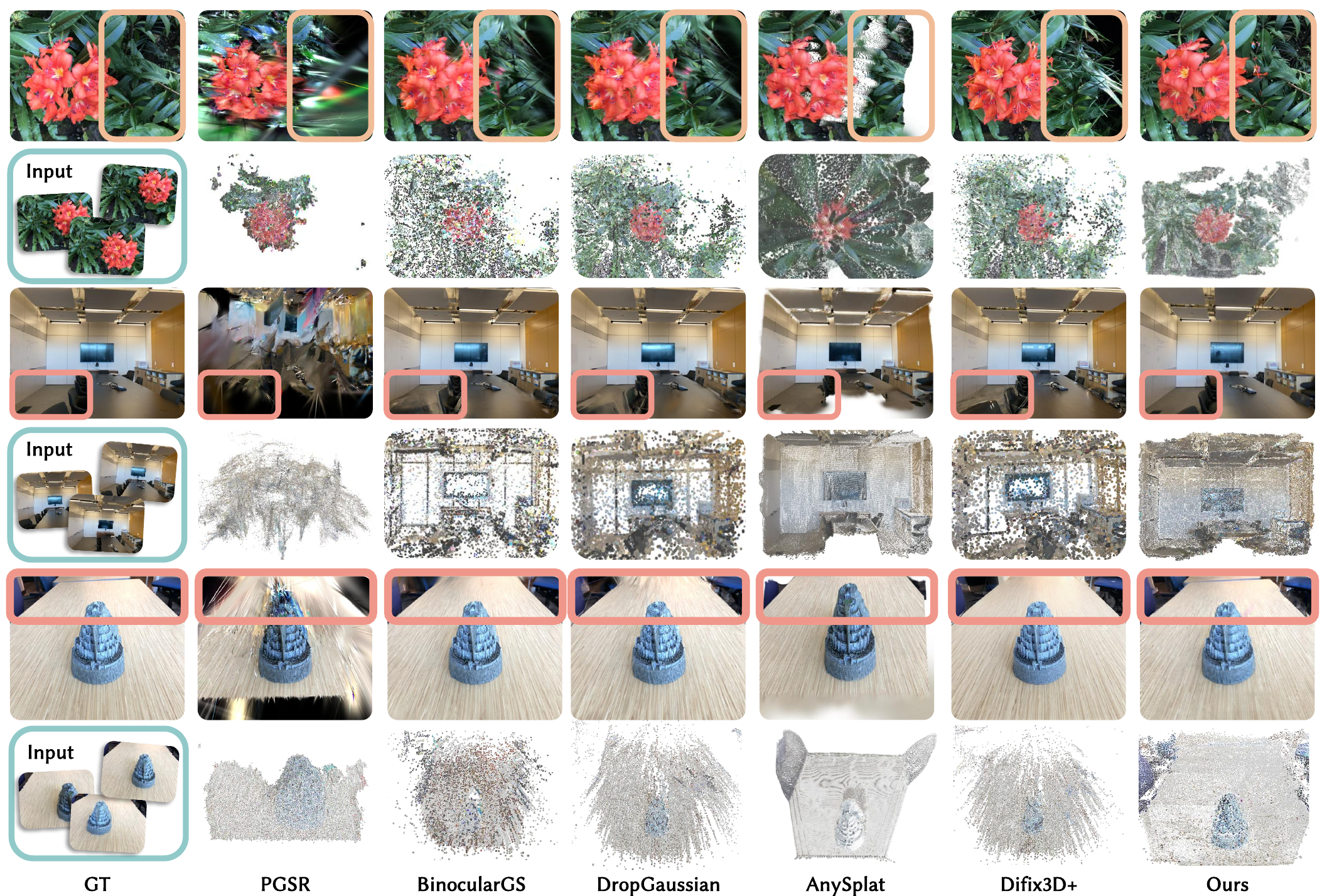}  
    \caption{Qualitative comparison on LLFF with three input views. We provide input images, novel-view renderings, and Gaussian positions. Other methods miss structures in unobserved areas and exhibit noisier Gaussian positions. In contrast, our co-optimization of geometry and generation reconstructs view-consistent novel views and complete scene geometry.
    }

    \label{fig:llff}
\end{figure*}

\begin{figure*}[h]
    \centering
    \includegraphics[width=1.0\linewidth]{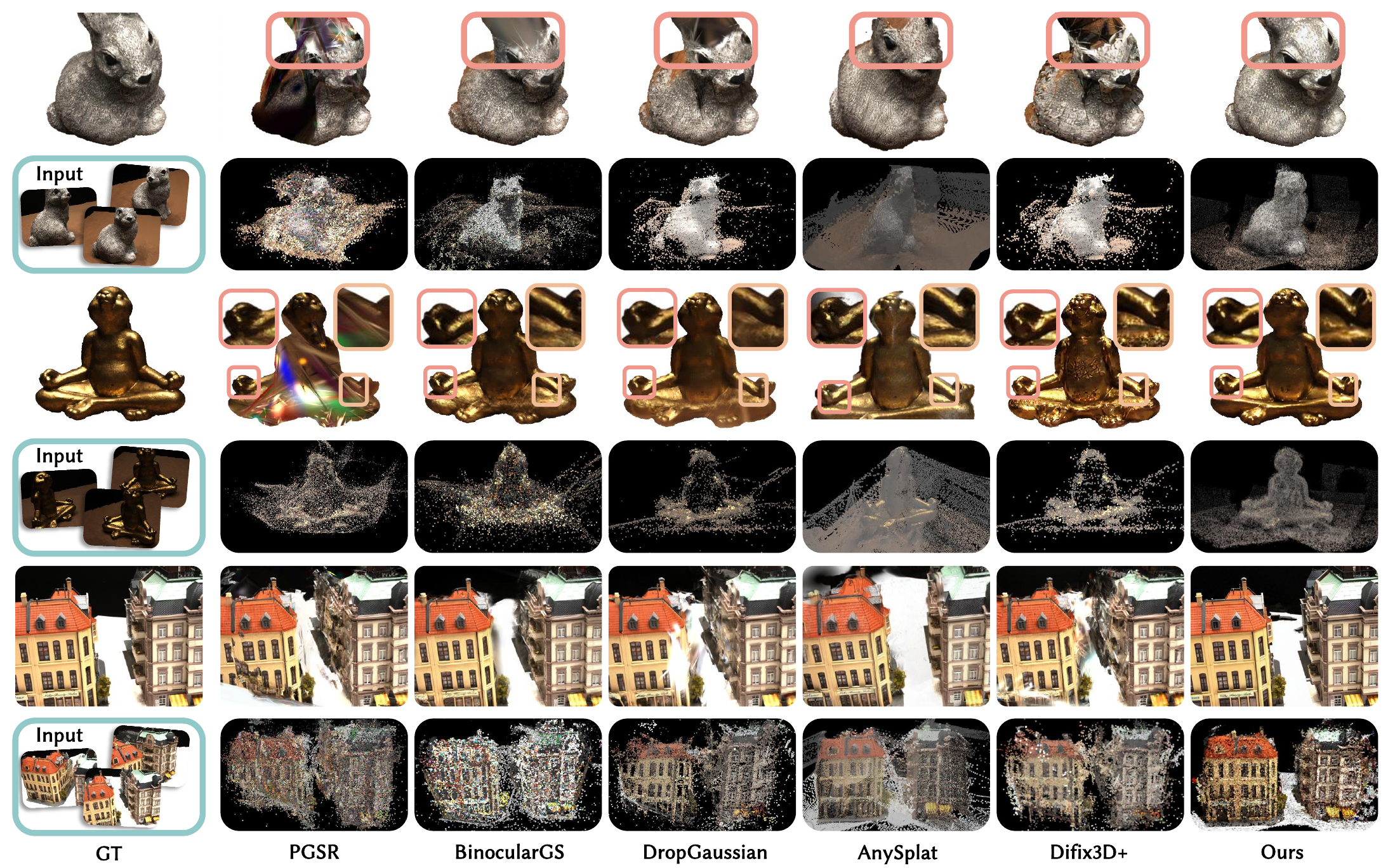}
    \caption{Qualitative comparison on DTU with three input views. We provide input images, novel-view renderings, and Gaussian positions. Other methods miss structures in unobserved regions (1st row), struggle to recover fine details (3rd and 5th rows), and exhibit noisier Gaussian positions. Our method reconstructs view-consistent, complete scene geometry while significantly improving novel-view visual quality.}
    \label{fig:dtu}
\end{figure*}

\begin{figure*}[h]
    \centering
    \includegraphics[width=1.0\linewidth]{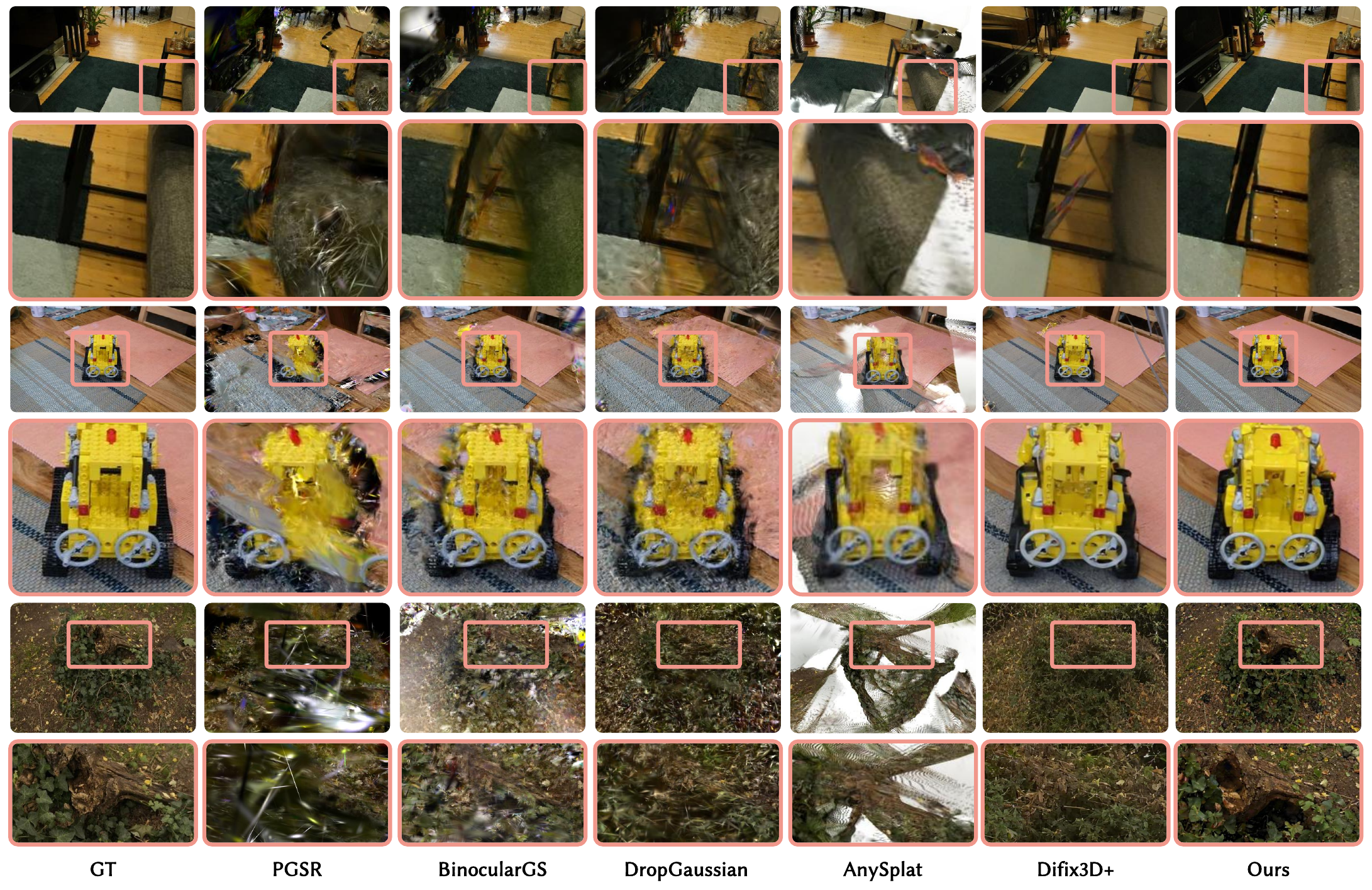}
    \caption{Qualitative comparison on Mip-NeRF 360 with three input views. For complex scenes, other methods suffer from degraded details when rendering novel views, whereas our method reconstructs fine-grained details with high fidelity.
}
    \label{fig:mipnerf}
\end{figure*}

\section{Experimental Results}
\subsection{Implementation Details}
We train for 15,000 iterations on the LLFF, DTU, and Mip-NeRF
360 datasets using an NVIDIA RTX 3090 GPU. The loss coefficients are empirically set as $\gamma_{\text{iso}}= 1.0$, $\gamma_{\text{bar}}=0.01$ (Eq.~\ref{loss_scale}), and $\gamma_{\text{i}}= 1.0$, $\gamma_{\text{n}}=0.5$, $\gamma_{\text{d}}=0.5$, $\gamma_{\text{e}}=0.01$, $\gamma_{\text{dif}}=0.1$ (Eq.~\ref{eq:total_loss}). The iterative optimization schedule is controlled by $T_1=3000$ and $T_2=100$.
\subsection{Datasets and Metrics}
\noindent\textbf{Datasets.} We conduct experiments on three public datasets, including the LLFF dataset \cite{llff}, the DTU dataset~\cite{dtu}, and Mip-NeRF 360 dataset~\cite{ReconFusion}. 
Following prior works \cite{Binoculargs,corgs,GenFusion}, we train with 3, 6, and 9 input views, and use the same test split as prior works for fair comparison.
The downsampling rates for the LLFF, DTU, and Mip-NeRF 360 datasets are 8, 4, and 4, respectively. 

\noindent\textbf{Metrics.} For quantitative evaluation, we adopt three widely used image quality metrics: Peak Signal-to-Noise Ratio~(PSNR), Structural Similarity Index~(SSIM)\cite{ssim}, and Learned Perceptual Image Patch Similarity~(LPIPS)\cite{lpips}. 
Specifically, PSNR measures the average peak error between the rendered images and the ground truth. SSIM evaluates structural similarity by considering luminance, contrast, and texture information. LPIPS computes perceptual distances using learned features.

\subsection{Baseline} Our comparison encompasses optimization-based methods, including FSGS~\cite{fsgs}, DNGaussian~\cite{dngaussian}, BinocularGS~\cite{Binoculargs}, CoR-GS~\cite{corgs}, DropGaussian~\cite{DropGaussian}, and the diffusion-guided Difix3D+~\cite{difix3d}, alongside the feed-forward approach AnySplat~\cite{anysplat}. The original PGSR~\cite{pgsr} is included as a baseline.

\subsection{Comparisons}
We compare our method with existing approaches through both quantitative and qualitative evaluations. To ensure fairness, all datasets follow the same experimental settings as previous works~\cite{Binoculargs,corgs,GenFusion}.

\noindent\textbf{LLFF.}
Table \ref{tab:llff_dtu_unified} shows the quantitative results of the LLFF dataset with 3, 6, and 9 input views, respectively. Our method outperforms all baselines in PSNR, SSIM, and LPIPS, with the largest improvement in LPIPS. 
Although CoR-GS\cite{corgs}, DropGaussian\cite{DropGaussian}, and BinocularGS\cite{Binoculargs} achieve competitive overall scores and improve as views increase, a consistent gap remains in perceptual quality as measured by LPIPS. 
DNGaussian~\cite{dngaussian} shows only slight improvement with more views, suggesting limited benefit from additional supervision due to errors and scale ambiguity in the monocular depth prior.
Difix3D+~\cite{difix3d} performs generative optimization on sparse-view reconstructions, but its improvements are limited by weak reconstructions and multi-view inconsistencies.

Fig.~\ref{fig:llff} presents the visual comparison of
novel view synthesis and Gaussian positions.
In the novel views, BinocularGS~\cite{Binoculargs} and DropGaussian~\cite{DropGaussian} reconstruct the visible regions reasonably well but suffer from missing geometry in unseen areas.
Difix3D+~\cite{difix3d} performs generative refinement, yielding slight appearance improvements but overall limited improvements. 
In the visualizations of Gaussian primitive positions, BinocularGS~\cite{Binoculargs}, DropGaussian~\cite{DropGaussian}, and Difix3D+~\cite{difix3d} produce noisy results with floaters, whereas AnySplat~\cite{anysplat} exhibits depth errors. Our method generates more complete, sharper novel views and more reasonable Gaussian positions with fewer artifacts.

\begin{table*}[ht]
\centering
\caption{Quantitative comparison on Mip-NeRF 360. 
The top-3 results in each column are marked as 
\textbf{\colorbox{tabfirst}{1st}, \colorbox{tabsecond}{2nd}, \colorbox{tabthird}{3rd}}.}
\label{tab:mipnerf360}
\scriptsize

\begin{tabular}{l|ccc|ccc|ccc}
\toprule
\multirow{3}{*}{\textbf{Methods}} &
\multicolumn{9}{c}{\textbf{Mip-NeRF 360}} \\
\cmidrule(lr){2-10}

& \multicolumn{3}{c|}{\textbf{3-view}} 
& \multicolumn{3}{c|}{\textbf{6-view}} 
& \multicolumn{3}{c}{\textbf{9-view}} \\

\cmidrule(lr){2-4}\cmidrule(lr){5-7}\cmidrule(lr){8-10}

& PSNR$\uparrow$ & SSIM$\uparrow$ & LPIPS$\downarrow$
& PSNR$\uparrow$ & SSIM$\uparrow$ & LPIPS$\downarrow$
& PSNR$\uparrow$ & SSIM$\uparrow$ & LPIPS$\downarrow$ \\

\midrule

PGSR~\cite{pgsr} & 11.57 & 0.193 & 0.666 & 13.92 & 0.301 & 0.546 & 15.09 & 0.367 & 0.476 \\

FSGS~\cite{fsgs} & 14.17 & 0.318 & 0.578 & 16.12 & 0.415 & 0.517 & \colorbox{tabthird}{17.94} & \colorbox{tabthird}{0.492} & 0.468 \\

DropGaussian~\cite{DropGaussian} & 14.08 & 0.325 & 0.609 & 16.11 & 0.406 & 0.535 & 17.46 & 0.473 & 0.489 \\

BinocularGS~\cite{Binoculargs} & 13.70 & \colorbox{tabthird}{0.361} & \colorbox{tabthird}{0.543} & 15.43 & \colorbox{tabthird}{0.427} & \colorbox{tabthird}{0.430} & 17.47 & \colorbox{tabsecond}{0.523} & \colorbox{tabthird}{0.341} \\

AnySplat~\cite{anysplat} & 8.91 & 0.165 & 0.659 & 10.89 & 0.197 & 0.596 & 11.43 & 0.204 & 0.580 \\

GenFusion~\cite{GenFusion} & \colorbox{tabsecond}{15.29} & \colorbox{tabsecond}{0.367} & 0.585 & \colorbox{tabsecond}{17.16} & \colorbox{tabsecond}{0.447} & 0.500 & \colorbox{tabsecond}{18.36} & 0.496 & 0.465 \\

DiFix3D+~\cite{difix3d} & \colorbox{tabthird}{14.63} & 0.317 & \colorbox{tabsecond}{0.487} & \colorbox{tabthird}{16.44} & 0.401 & \colorbox{tabsecond}{0.391} & 17.76 & 0.461 & \colorbox{tabfirst}{0.335} \\

Ours & \colorbox{tabfirst}{17.69} & \colorbox{tabfirst}{0.538} & \colorbox{tabfirst}{0.442} & \colorbox{tabfirst}{18.80} & \colorbox{tabfirst}{0.608} & \colorbox{tabfirst}{0.394} & \colorbox{tabfirst}{20.57} & \colorbox{tabfirst}{0.723} & \colorbox{tabsecond}{0.338} \\

\bottomrule
\end{tabular}

\end{table*}

\noindent\textbf{DTU.}
Table \ref{tab:llff_dtu_unified} shows the quantitative results of the DTU dataset under 3, 6, and 9 input views, respectively.
Our method performs best with 3 and 6 input views. At 9 views, CoR-GS~\cite{corgs} slightly leads in PSNR (0.02 dB), whereas we achieve higher SSIM and lower LPIPS.
Notably, with 6 input views, our method achieves the largest gains over the second best, improving PSNR by 1.53 dB, SSIM by 0.019, and reducing LPIPS by 50\%, demonstrating stable geometry and accurate fine-detail recovery under uniform view coverage.
Fig.~\ref{fig:dtu} compares novel-view renderings and Gaussian positions.  Other methods miss structures in unobserved areas, struggle to recover fine details, and exhibit noisier Gaussian positions.

\noindent\textbf{Mip-NeRF 360.} Table~\ref{tab:mipnerf360} shows the quantitative results on the Mip-NeRF 360 dataset, which features complex scene structures and large viewpoint variations. Our method achieves the best performance under the 3-view setting across all metrics. Specifically, it surpasses the second-best method by 2.40 dB in PSNR, 0.171 in SSIM, and reduces LPIPS by 0.045, demonstrating a clear advantage in recovering both geometric structures and fine-grained details under extremely sparse inputs. 
Fig.~\ref{fig:mipnerf} shows that other methods tend to suffer from noticeable detail degradation when rendering novel views in complex scenes, often leading to blurry textures or missing structures. In contrast, our method preserves rich high-frequency details and maintains structural consistency across views, producing more complete and visually faithful novel view renderings without the common issue of detail loss.

\subsection{Ablation Study}

To comprehensively evaluate our design, we conduct a series of ablation and analysis experiments covering both core components and key design choices. Specifically, we study the effects of metric depth, DINO features, and the diffusion module, as well as the impact of diffusion variants and initialization strategies. All experiments are conducted on representative subsets of the DTU, LLFF, and Mip-NeRF 360 datasets under a 3-view setting for efficient ablation analysis.

\begin{table}[ht!]
\footnotesize
\caption{Ablation study on the proposed components.
We evaluate metric depth, DINO features, and diffusion refinement on LLFF and DTU. The first four rows analyze component contributions, while the last four rows show that removing any module degrades performance.}
\centering
\setlength{\tabcolsep}{3pt}
\resizebox{\linewidth}{!}{
\begin{tabular}{@{}ccc|ccc|ccc@{}}
 \toprule
\multirow{2}{*}{Depth} &
\multirow{2}{*}{DINO} &
\multirow{2}{*}{Diffusion} &
\multicolumn{3}{c|}{\textbf{DTU}} &
\multicolumn{3}{c}{\textbf{LLFF}} \\
 & & & PSNR$\uparrow$ & SSIM$\uparrow$ & LPIPS$\downarrow$
            & PSNR$\uparrow$ & SSIM$\uparrow$ & LPIPS$\downarrow$ \\  
\midrule
 &  &   & 16.89&0.770&0.181&15.96&0.624&0.303\\
\cmark  &&& 19.14 & 0.790 & 0.120 & 17.40 & 0.631 & 0.224 \\
& \cmark  && 18.03 & 0.786 & 0.120 & 17.11 & 0.670 & 0.224 \\
&& \cmark & 17.91 & 0.782 & 0.121 & 17.08 & 0.673 & 0.224 \\
\midrule
 & \cmark & \cmark  & 19.01&0.853&0.111&18.34&0.726&0.212 \\
\cmark &  & \cmark  &  19.55&0.852&0.110&18.61& 0.676&0.203\\
\cmark & \cmark &   &  19.57&0.851& 0.109&18.40&0.649& 0.211\\
\cmark & \cmark & \cmark  & \textbf{20.75} & \textbf{0.873} & \textbf{0.091} & \textbf{20.73} & \textbf{0.761} & \textbf{0.168} \\
\bottomrule
\end{tabular}
}

\label{table:ablation_main}
\end{table}

\subsubsection{Effectiveness of Metric Depth}

\begin{figure}[h]

    \centering
    \includegraphics[width=\linewidth]{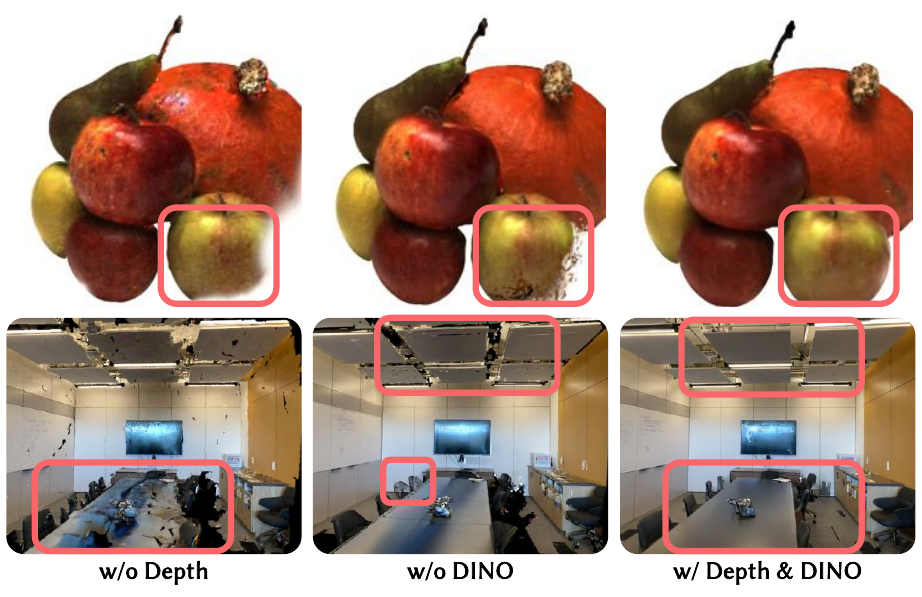}  

    \caption{Ablation Study by Visualization. Without metric depth (1st column), structural errors appear, and without DINO features (2nd column), the results are structurally incomplete.
    }
    \label{fig:abl1}
  
\end{figure}

To verify the effectiveness of metric depth, we compare results with and without this component in Table~\ref{table:ablation_main}. Introducing metric depth (row 2) significantly improves reconstruction quality, boosting PSNR from 16.89 dB to 19.14 dB on DTU and from 15.96 dB to 17.40 dB on LLFF. This demonstrates that metric depth provides reliable geometric guidance, leading to more accurate and stable reconstruction.
As shown in Fig.~\ref{fig:abl1}, the absence of scale-consistent depth leads to unstable geometry estimation and distorted surface structures. This lack of accurate geometric grounding in turn causes the diffusion module to amplify these errors during refinement, resulting in structural artifacts and degraded novel-view rendering quality.

\subsubsection{Effectiveness of Our Metric Depth Estimation}
\begin{table}[h]
\centering
\small
\caption{Comparison of different metric depth estimation methods on the Mip-NeRF 360 dataset, where all methods are evaluated under the same optimization pipeline.}
\setlength{\tabcolsep}{6pt}
\begin{tabular}{l|ccc}
\toprule
\textbf{Metric Depth Methods} & \textbf{PSNR} $\uparrow$ & \textbf{SSIM} $\uparrow$ & \textbf{LPIPS} $\downarrow$ \\
\midrule

DepthAnythingV2~\cite{depthanythingV2}& 12.62 &  0.284 &0.633\\
Ours &\textbf{17.11}&\textbf{0.506}&\textbf{0.480}\\
\bottomrule
\end{tabular}

\label{tab:metric_depth_ablation}
\end{table}
\begin{figure}[h]
    \centering
    \includegraphics[width=\linewidth]{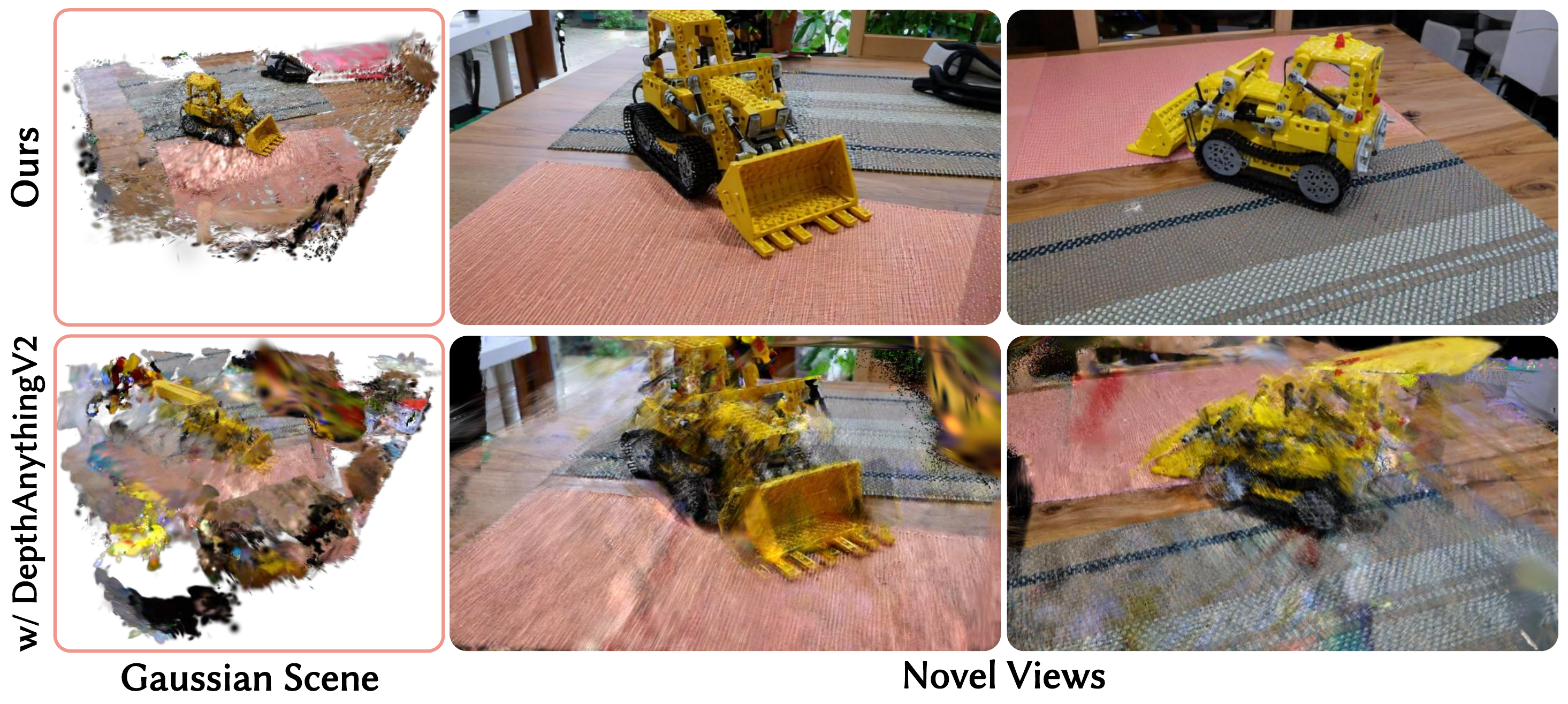}  
    \caption{Visualization of Gaussian scene reconstruction and novel-view rendering under different metric depth inputs. Results using metric depth estimated by DepthAnythingV2~\cite{depthanythingV2} exhibit more floaters in the scene and blur in novel views, while our method yields cleaner and more stable reconstructions.
    }
    \label{fig:da2}
\end{figure}

To validate the effectiveness of our metric depth estimation, we compare it with metric depth predicted by DepthAnythingV2~\cite{depthanythingV2}. Both depth sources are integrated into the same Gaussian optimization pipeline, with all other settings kept identical. Quantitative results are reported in Table~\ref{tab:metric_depth_ablation}. On the Mip-NeRF 360 dataset, our method significantly outperforms DepthAnythingV2 across all metrics, improving PSNR by 4.49 dB, SSIM by 0.222, and reducing LPIPS by 24.2\%. This demonstrates that our method provides more accurate and consistent geometric constraints, leading to improved reconstruction quality in the Gaussian representation.
As shown in Fig.~\ref{fig:da2}, directly using metric depth estimated by DepthAnythingV2~\cite{depthanythingV2} leads to unstable geometry, where the reconstructed Gaussian scene contains a significant number of floaters and noisy artifacts. These geometric inconsistencies further propagate to novel-view rendering, causing noticeable blur and degraded visual fidelity.
In contrast, the proposed metric depth provides more accurate and consistent geometric guidance during optimization. This results in a cleaner and more stable Gaussian scene representation, effectively reducing floaters. Consequently, the rendered novel views exhibit sharper details and improved structural consistency.

\subsubsection{Effectiveness of DINO Feature}

To verify the effectiveness of the DINO feature, we compare the results with and without this module. As shown in Fig.~\ref{fig:abl1}, removing DINO leads to incomplete reconstruction, especially for isolated or thin objects such as the apple and the chair. Quantitatively, as shown in Tab.~\ref{table:ablation_main}, introducing DINO significantly improves reconstruction quality (row 3), increasing PSNR from 16.89 dB to 18.03 dB on DTU and from 15.96 dB to 17.11 dB on LLFF. 
Furthermore, in the combined settings, removing DINO (row 6) leads to a noticeable performance drop compared to the full model (row 8), with PSNR decreasing from 20.75 dB to 19.55 dB on DTU and from 20.73 dB to 18.61 dB on LLFF, indicating that DINO remains crucial across different module combinations.

Additionally, we assess the impact of feature dimensionality on the DTU and LLFF datasets by comparing PCA-reduced features at 32 and 3 dimensions, as shown in Table \ref{table:ablation_dim}. The 3D variant performs better, which demonstrates that low-frequency features are more effective in maintaining structural consistency.

\begin{table}[h]
\scriptsize
\caption{Ablation study on DINO feature extraction strategies.
3D PCA-reduced DINO features outperform 32D, showing stronger low-frequency structural consistency.}
\setlength{\tabcolsep}{5pt}
\centering
\begin{tabular}{l|ccc|ccc}
\toprule
\multirow{2}{*}{\textbf{Feature} }
 & \multicolumn{3}{c|}{\textbf{DTU}} 
 & \multicolumn{3}{c}{\textbf{LLFF}} \\ 
\cmidrule(lr){2-7}
 \textbf{Dimension}& PSNR$\uparrow$ & SSIM$\uparrow$ & LPIPS$\downarrow$ 
 & PSNR$\uparrow$ & SSIM$\uparrow$ & LPIPS$\downarrow$ \\  
\midrule
32D         &  19.72&0.851 &0.112&18.38&0.718&0.195\\
3D       & \textbf{20.75} &\textbf{0.873}&\textbf{0.091}  &\textbf{20.73}&\textbf{0.761}&\textbf{0.168}\\
\bottomrule
\end{tabular}

\label{table:ablation_dim}
\end{table}
\subsubsection{Effectiveness of RGB Diffusion} 
\begin{figure}[h]

    \centering
    \includegraphics[width=1.0\linewidth]{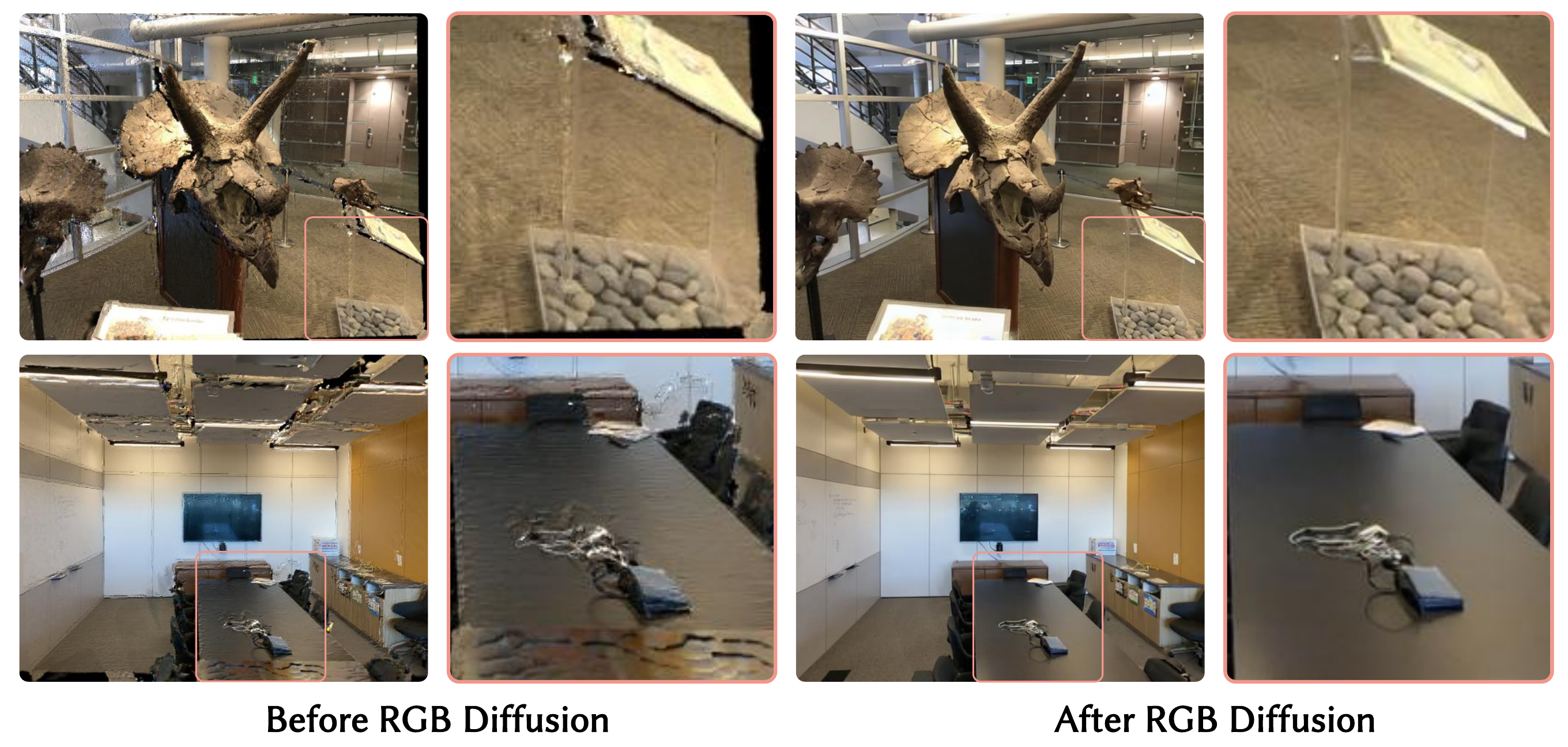}  

    \caption{Visual comparison before and after RGB diffusion refinement. Without RGB diffusion, the rendered images contain missing regions and limited details. After applying RGB diffusion, finer details are recovered and previously missing structures are completed, resulting in improved visual fidelity.
    }
    \label{fig:abl4}
  
\end{figure}
To validate the effectiveness of the diffusion module, we conduct both image-level and optimization-level comparisons. As shown in Fig.~\ref{fig:abl4}, we first compare the rendered images before and after applying the diffusion module, where diffusion enhances high-frequency details and improves visual fidelity. Furthermore, Fig.~\ref{fig:abl2} presents a comparison within the Gaussian optimization process, where we evaluate results with and without the novel-view diffusion module. Removing the diffusion module leads to incomplete scene structures and noticeable loss of fine details, while incorporating it produces more coherent and detailed reconstructions.

Quantitatively, Tab.~\ref{table:ablation_main} shows that introducing diffusion improves perceptual quality (row 4), reducing LPIPS from 0.181 to 0.121 on DTU and from 0.303 to 0.224 on LLFF. 
Furthermore, in the combined settings, removing diffusion (row 7) degrades performance compared to the full model (row 8), with LPIPS increasing from 0.091 to 0.109 on DTU and from 0.168 to 0.211 on LLFF, indicating that diffusion plays a crucial role in enhancing fine details and visual realism.

\subsubsection{Effectiveness of Iterative Diffusion} 
\begin{table}[h]
\scriptsize
\caption{ Ablation study on RGB Diffusion module strategies.
We compare post-training (w/o iterative) and iterative diffusion refinements, finding that the iterative setting yields the best overall results. }
\centering
\setlength{\tabcolsep}{5pt}
\begin{tabular}{l|ccc|ccc}
\toprule
\multirow{2}{*}{\textbf{Diffusion }} 
 & \multicolumn{3}{c|}{\textbf{DTU}} 
 & \multicolumn{3}{c}{\textbf{LLFF}} \\ 
\cmidrule(lr){2-7}
\textbf{Refinement} & PSNR$\uparrow$ & SSIM$\uparrow$ & LPIPS$\downarrow$ 
 & PSNR$\uparrow$ & SSIM$\uparrow$ & LPIPS$\downarrow$ \\  
\midrule
w/o Iterative & 19.34&0.845&0.110&18.70&0.676&0.185 \\
w/ Iterative  & \textbf{20.75} & \textbf{0.873} & \textbf{0.091} & \textbf{20.73} & \textbf{0.761} & \textbf{0.168} \\
\bottomrule

\end{tabular}

\
\label{table:ablation_Difix}
\end{table}

\begin{figure}[h]
    \centering
    \includegraphics[width=1.0\linewidth]{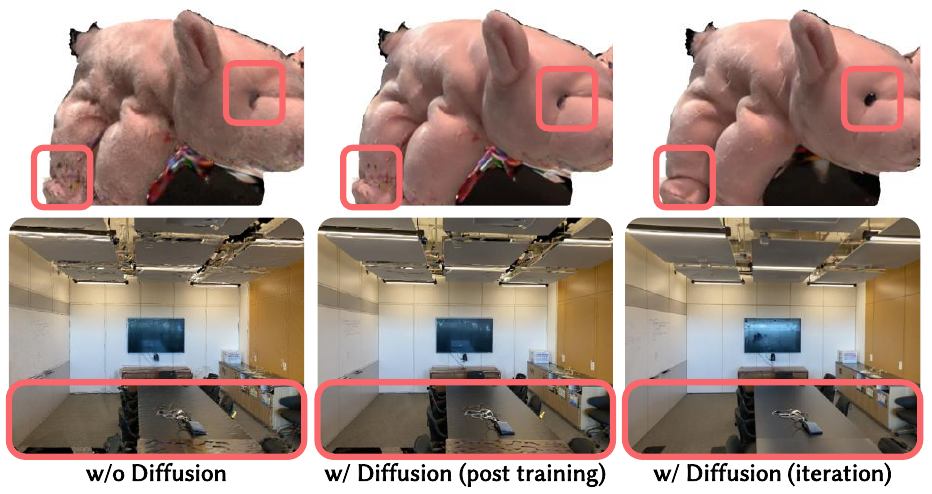}  
 
    \caption{Ablation Study by Visualization. Without diffusion, fine details remain incomplete. While post-training (w/o iterative) refinement offers some improvement, its capabilities are limited. In contrast, iterative diffusion consistently recovers fine details.
    }
    \label{fig:abl2}
\end{figure}

We further evaluate different diffusion strategies by comparing iterative and post-training variants in Tab.~\ref{table:ablation_Difix}. The iterative setting consistently achieves superior performance across all metrics. Specifically, it improves PSNR from 19.34 to 20.75 on DTU and from 18.70 to 20.73 on LLFF, while also reducing LPIPS from 0.110 to 0.091 and from 0.185 to 0.168, respectively. 
Fig.~\ref{fig:abl2} illustrates that the iterative Diffusion recovers richer and more stable details than the post-training strategy.

\subsubsection{Analysis of Fine-Tuned Diffusion Model}
\begin{table}[h]
\scriptsize
\caption{Comparative results with and without fine-tuning the RGB Diffusion Model. The data demonstrate the significant performance improvement achieved by utilizing the model that has been fine-tuned. }
\centering
\setlength{\tabcolsep}{5pt}
\begin{tabular}{l|ccc|ccc}
\toprule
\multirow{2}{*}{\textbf{Diffusion }} 
 & \multicolumn{3}{c|}{\textbf{DTU}} 
 & \multicolumn{3}{c}{\textbf{LLFF}} \\ 
\cmidrule(lr){2-7}
\textbf{Model} & PSNR$\uparrow$ & SSIM$\uparrow$ & LPIPS$\downarrow$ 
 & PSNR$\uparrow$ & SSIM$\uparrow$ & LPIPS$\downarrow$ \\  
\midrule
w/o fintune &19.76 &0.816 &0.096 &19.78 &0.755& 0.176 \\
w/ finetune  & \textbf{20.75} & \textbf{0.873} & \textbf{0.091} & \textbf{20.73} & \textbf{0.761} & \textbf{0.168} \\
\bottomrule
\end{tabular}
\label{abl:finetune}
\end{table}
As shown in Table~\ref{abl:finetune}, we conducted a comparative analysis on the LLFF and DTU datasets to evaluate the effect of fine-tuning the RGB Diffusion Model. We compared the iterative refinement results utilizing an RGB Diffusion Model fine-tuned on SynCamMaster datasets~\cite{syncammaster} against the non-fine-tuned baseline results. The quantitative analysis clearly demonstrates that the fine-tuned model achieves a significant performance improvement, consistently yielding the best overall performance metrics.

\subsubsection{Comparison with Diffusion-based Methods} 
\begin{table}[h]
\centering
\small
\caption{Comparison of different diffusion backbones on the LLFF dataset. All methods share the same Gaussian initialization and 3DGS optimization pipeline.}
\setlength{\tabcolsep}{6pt}
\begin{tabular}{l|ccc}
\toprule
\textbf{Diffusion Backbones} & \textbf{PSNR} $\uparrow$ & \textbf{SSIM} $\uparrow$ & \textbf{LPIPS} $\downarrow$ \\
\midrule

G4Splat~\cite{ni2025g4splat}&  18.02& 0.642&0.326\\
GenFusion~\cite{GenFusion} &  17.70& 0.609 &0.245  \\
Ours &\textbf{20.73}&\textbf{0.761}&\textbf{0.168}\\
\bottomrule
\end{tabular}

\label{tab:diffusion_ablation}
\end{table}
\begin{figure}[h]
    \centering
    \includegraphics[width=1.0\linewidth]{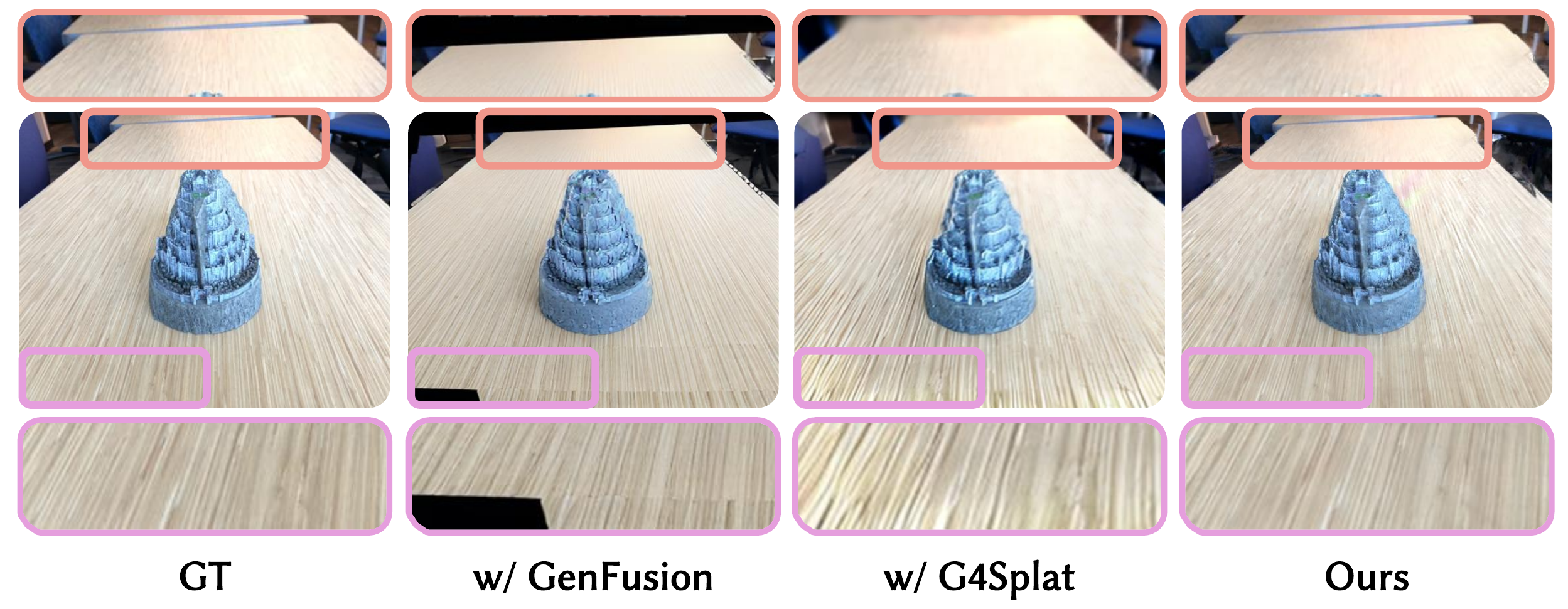}  
 
    \caption{Qualitative comparison of different diffusion backbones on LLFF. Compared to G4Splat~\cite{ni2025g4splat} and GenFusion~\cite{GenFusion}, our diffusion method produces richer details and clearer structures.
    }
    \label{fig:dif_dif}
\end{figure}
To evaluate the impact of different diffusion models, we replace the diffusion module in our method with alternative backbones, including G4Splat~\cite{ni2025g4splat} and GenFusion~\cite{GenFusion}, while keeping the same Gaussian initialization and optimization pipeline. All experiments are conducted on the LLFF dataset. As shown in As shown in Table~\ref{tab:diffusion_ablation}, our method achieves the best performance across all metrics, reaching a PSNR of 20.73 dB and outperforming G4Splat at 18.02 dB and GenFusion at 17.70 dB. From Fig.~\ref{fig:dif_dif}, our method produces more detailed textures and clearer structures, while the compared methods suffer from blur or structural inconsistencies.

\subsubsection{Analysis of Initialization Strategy}
\begin{table}[h]
\centering
\small
\caption{Ablation on Gaussian initialization. We replace our depth back-projected initialization with VGGT point clouds and evaluate on LLFF, demonstrating consistent performance and robustness to different initializations.}
\setlength{\tabcolsep}{6pt}
\begin{tabular}{l|ccc}
\toprule
\textbf{Init} & \textbf{PSNR} $\uparrow$ & \textbf{SSIM} $\uparrow$ & \textbf{LPIPS} $\downarrow$ \\
\midrule
 VGGT~\cite{wang2025vggt} & 20.63  & 0.756  & 0.176 \\
Ours &\textbf{20.73}&\textbf{0.761}&\textbf{0.168}\\
\bottomrule
\end{tabular}

\label{tab:init_ablation}
\end{table}
Gaussian initialization in our framework is primarily an implementation choice rather than a core contribution. By default, we initialize the Gaussian primitives using a point cloud obtained via depth back-projection, which provides a reasonable geometric prior for subsequent optimization. To evaluate the sensitivity to initialization, we replace it with VGGT~\cite{wang2025vggt} point clouds and conduct experiments on the LLFF dataset with 3 input views. As shown in Table~\ref{tab:init_ablation}, the performance remains comparable, demonstrating that our method is robust to different initialization strategies.
\section{Conclusion}
We propose D$^{3}$GS, a Depth–DINO–Diffusion guided framework for sparse-view 3D Gaussian Splatting that addresses the lack of geometric and photometric supervision. 
By injecting metric depth, DINO-based view-consistent feature, and diffusion-driven novel-view refinement directly into the Gaussian representation, D$^{3}$GS jointly enhances geometric accuracy and appearance fidelity under highly sparse inputs. 
It employs metric depth completion with a denoising U-Net and DPT refinement to provide robust geometric initialization. DINO-guided view-consistent feature learning is further introduced to enhance structural consistency across views. Finally, a diffusion-driven iterative refinement module recovers high-frequency details. 
Extensive experiments show that our method achieves substantial and consistent improvements over state-of-the-art baselines on standard datasets, validating the effectiveness and complementarity of the three guidance components.

Despite these advantages, our method introduces additional overhead from depth estimation, feature extraction, and the diffusion model, resulting in a moderate increase in overall processing and training time (approximately 25 minutes per scene). Future work will focus on improving efficiency by reducing computational overhead and extending the framework to dynamic scenes through temporal modeling.

\section*{Acknowledgments}
This work was supported in part by the National Natural Science Foundation of China under Grants 62077026, 62125107, and 62402274, and by the National Key Research and Development Program of China under Grant 2024YFC3308300.



 
%

\bibliographystyle{IEEEtran}
\bibliography{sample-bibliography.bib}











\vfill

\end{document}